\documentclass{article} 
\usepackage{iclr2027_conference,times}

\usepackage{amsmath,amsfonts,bm}

\def\eqref#1{equation~\ref{#1}}

\def\1{\bm{1}}

\DeclareMathAlphabet{\mathsfit}{\encodingdefault}{\sfdefault}{m}{sl}
\SetMathAlphabet{\mathsfit}{bold}{\encodingdefault}{\sfdefault}{bx}{n}

\usepackage{soul}
\usepackage{url}  
\usepackage{microtype}
\usepackage{booktabs} 
\usepackage{hyperref}
\usepackage{listings}
\usepackage{makecell}
\usepackage{tcolorbox}
\usepackage{enumitem}
\usepackage{subfig}
\usepackage{float}
\usepackage{placeins}
\usepackage{wrapfig}
\usepackage{needspace}
\usepackage{multirow} 
\usepackage{caption}
\usepackage{graphicx}
\usepackage{subcaption}
\usepackage{color}
\usepackage{algorithm}
\usepackage{algpseudocode}
\usepackage{tcolorbox}
\definecolor{keywordcolor}{rgb}{0.7, 0.1, 0.1}   
\definecolor{tacticcolor}{rgb}{0.0, 0.1, 0.6}    
\definecolor{commentcolor}{rgb}{0.4, 0.4, 0.4}   
\definecolor{symbolcolor}{rgb}{0.0, 0.1, 0.6}    
\definecolor{sortcolor}{rgb}{0.1, 0.5, 0.1}      
\definecolor{attributecolor}{rgb}{0.7, 0.1, 0.1} 
\colorlet{qmapA}{blue!14}
\colorlet{qmapB}{green!20}
\newcommand{\qA}[1]{{\sethlcolor{qmapA}\hl{#1}}}
\newcommand{\qB}[1]{{\sethlcolor{qmapB}\hl{#1}}}

\begingroup
\lstdefinelanguage{lean} {

mathescape=false,
texcl=false,

morekeywords=[1]{
import, prelude, protected, private, noncomputable, definition, meta, renaming,
hiding, parameter, parameters, begin, constant, constants,
lemma, variable, variables, theory,
print, theorem, example,
open, as, export, override, axiom, axioms, inductive, with,
structure, record, universe, universes,
alias, help, precedence, reserve, declare_trace, add_key_equivalence,
match, infix, infixl, infixr, notation, postfix, prefix, instance,
eval, reduce, check, end, this,
using, using_well_founded, namespace, section,
attribute, local, set_option, extends, include, omit, class,
raw, replacing,
calc, have, show, suffices, by, in, at, let, forall, Pi, fun,
exists, if, dif, then, else, assume, obtain, from, register_simp_ext, unless, break, continue,
mutual, do, def, run_cmd, const,
partial, mut, where, macro, syntax, deriving,
return, try, catch, for, macro_rules, declare_syntax_cat, abbrev},

morekeywords=[2]{Sort, Type, Prop},

morekeywords=[3]{
assumption,
apply, intro, intros, allGoals,
generalize, clear, revert, done, exact,
refine, repeat, cases, rewrite, rw,
simp, simp_all, contradiction,
constructor, injection,
induction,
},

literate=
{α}{{\ensuremath{\mathrm{\alpha}}}}1
{β}{{\ensuremath{\mathrm{\beta}}}}1
{γ}{{\ensuremath{\mathrm{\gamma}}}}1
{δ}{{\ensuremath{\mathrm{\delta}}}}1
{ε}{{\ensuremath{\mathrm{\varepsilon}}}}1
{ζ}{{\ensuremath{\mathrm{\zeta}}}}1
{η}{{\ensuremath{\mathrm{\eta}}}}1
{θ}{{\ensuremath{\mathrm{\theta}}}}1
{ι}{{\ensuremath{\mathrm{\iota}}}}1
{κ}{{\ensuremath{\mathrm{\kappa}}}}1
{μ}{{\ensuremath{\mathrm{\mu}}}}1
{ν}{{\ensuremath{\mathrm{\nu}}}}1
{ξ}{{\ensuremath{\mathrm{\xi}}}}1
{π}{{\ensuremath{\mathrm{\mathnormal{\pi}}}}}1
{ρ}{{\ensuremath{\mathrm{\rho}}}}1
{σ}{{\ensuremath{\mathrm{\sigma}}}}1
{τ}{{\ensuremath{\mathrm{\tau}}}}1
{φ}{{\ensuremath{\mathrm{\varphi}}}}1
{χ}{{\ensuremath{\mathrm{\chi}}}}1
{ψ}{{\ensuremath{\mathrm{\psi}}}}1
{ω}{{\ensuremath{\mathrm{\omega}}}}1

{Γ}{{\ensuremath{\mathrm{\Gamma}}}}1
{Δ}{{\ensuremath{\mathrm{\Delta}}}}1
{Θ}{{\ensuremath{\mathrm{\Theta}}}}1
{Λ}{{\ensuremath{\mathrm{\Lambda}}}}1
{Σ}{{\ensuremath{\mathrm{\Sigma}}}}1
{Φ}{{\ensuremath{\mathrm{\Phi}}}}1
{Ξ}{{\ensuremath{\mathrm{\Xi}}}}1
{Ψ}{{\ensuremath{\mathrm{\Psi}}}}1
{Ω}{{\ensuremath{\mathrm{\Omega}}}}1

{ℵ}{{\ensuremath{\aleph}}}1

{≤}{{\ensuremath{\leq}}}1
{≥}{{\ensuremath{\geq}}}1
{≠}{{\ensuremath{\neq}}}1
{≈}{{\ensuremath{\approx}}}1
{≡}{{\ensuremath{\equiv}}}1
{≃}{{\ensuremath{\simeq}}}1

{≤}{{\ensuremath{\leq}}}1
{≥}{{\ensuremath{\geq}}}1

{∂}{{\ensuremath{\partial}}}1
{∆}{{\ensuremath{\triangle}}}1 

{∫}{{\ensuremath{\int}}}1
{∑}{{\ensuremath{\mathrm{\Sigma}}}}1
{Π}{{\ensuremath{\mathrm{\Pi}}}}1

{⊥}{{\ensuremath{\perp}}}1
{∞}{{\ensuremath{\infty}}}1
{∂}{{\ensuremath{\partial}}}1

{∓}{{\ensuremath{\mp}}}1
{±}{{\ensuremath{\pm}}}1
{×}{{\ensuremath{\times}}}1

{⊕}{{\ensuremath{\oplus}}}1
{⊗}{{\ensuremath{\otimes}}}1
{⊞}{{\ensuremath{\boxplus}}}1

{∇}{{\ensuremath{\nabla}}}1
{√}{{\ensuremath{\sqrt}}}1

{⬝}{{\ensuremath{\cdot}}}1
{•}{{\ensuremath{\cdot}}}1
{∘}{{\ensuremath{\circ}}}1

{⁻}{{\ensuremath{^{-}}}}1
{▸}{{\ensuremath{\blacktriangleright}}}1

{∧}{{\ensuremath{\wedge}}}1
{∨}{{\ensuremath{\vee}}}1
{¬}{{\ensuremath{\neg}}}1
{⊢}{{\ensuremath{\vdash}}}1

{⟨}{{\ensuremath{\langle}}}1
{⟩}{{\ensuremath{\rangle}}}1

{↦}{{\ensuremath{\mapsto}}}1
{←}{{\ensuremath{\leftarrow}}}1
{<-}{{\ensuremath{\leftarrow}}}1
{→}{{\ensuremath{\rightarrow}}}1
{↔}{{\ensuremath{\leftrightarrow}}}1
{⇒}{{\ensuremath{\Rightarrow}}}1
{⟹}{{\ensuremath{\Longrightarrow}}}1
{⇐}{{\ensuremath{\Leftarrow}}}1
{⟸}{{\ensuremath{\Longleftarrow}}}1

{∩}{{\ensuremath{\cap}}}1
{∪}{{\ensuremath{\cup}}}1
{⊂}{{\ensuremath{\subseteq}}}1
{⊆}{{\ensuremath{\subseteq}}}1
{⊄}{{\ensuremath{\nsubseteq}}}1
{⊈}{{\ensuremath{\nsubseteq}}}1
{⊃}{{\ensuremath{\supseteq}}}1
{⊇}{{\ensuremath{\supseteq}}}1
{⊅}{{\ensuremath{\nsupseteq}}}1
{⊉}{{\ensuremath{\nsupseteq}}}1
{∈}{{\ensuremath{\in}}}1
{∉}{{\ensuremath{\notin}}}1
{∋}{{\ensuremath{\ni}}}1
{∌}{{\ensuremath{\notni}}}1
{∅}{{\ensuremath{\emptyset}}}1

{∖}{{\ensuremath{\setminus}}}1
{†}{{\ensuremath{\dag}}}1

{ℕ}{{\ensuremath{\mathbb{N}}}}1
{ℤ}{{\ensuremath{\mathbb{Z}}}}1
{ℝ}{{\ensuremath{\mathbb{R}}}}1
{ℚ}{{\ensuremath{\mathbb{Q}}}}1
{ℂ}{{\ensuremath{\mathbb{C}}}}1
{⌞}{{\ensuremath{\llcorner}}}1
{⌟}{{\ensuremath{\lrcorner}}}1
{⦃}{{\ensuremath{\{\!|}}}1
{⦄}{{\ensuremath{|\!\}}}}1

{‖}{{\ensuremath{\|}}}1
{₁}{{\ensuremath{_1}}}1
{₂}{{\ensuremath{_2}}}1
{₃}{{\ensuremath{_3}}}1
{₄}{{\ensuremath{_4}}}1
{₅}{{\ensuremath{_5}}}1
{₆}{{\ensuremath{_6}}}1
{₇}{{\ensuremath{_7}}}1
{₈}{{\ensuremath{_8}}}1
{₉}{{\ensuremath{_9}}}1
{₀}{{\ensuremath{_0}}}1
{ᵢ}{{\ensuremath{_i}}}1
{ⱼ}{{\ensuremath{_j}}}1
{ₐ}{{\ensuremath{_a}}}1

{¹}{{\ensuremath{^1}}}1

{ₙ}{{\ensuremath{_n}}}1
{ₘ}{{\ensuremath{_m}}}1
{ₚ}{{\ensuremath{_p}}}1
{↑}{{\ensuremath{\uparrow}}}1
{↓}{{\ensuremath{\downarrow}}}1

{...}{{\ensuremath{\ldots}}}1
{·}{{\ensuremath{\cdot}}}1

{▸}{{\ensuremath{\triangleright}}}1

{Σ}{{\color{symbolcolor}\ensuremath{\Sigma}}}1
{Π}{{\color{symbolcolor}\ensuremath{\Pi}}}1
{∀}{{\color{symbolcolor}\ensuremath{\forall}}}1
{∃}{{\color{symbolcolor}\ensuremath{\exists}}}1
{λ}{{\color{symbolcolor}\ensuremath{\mathrm{\lambda}}}}1
{\$}{{\color{symbolcolor}\$}}1

{:=}{{\color{symbolcolor}:=}}1
{=}{{\color{symbolcolor}=}}1
{<|>}{{\color{symbolcolor}<|>}}1
{<\$>}{{\color{symbolcolor}<\$>}}1
{+}{{\color{symbolcolor}+}}1
{*}{{\color{symbolcolor}*}}1,

morecomment=[s][\color{commentcolor}]{/-}{-/},
morecomment=[l][\itshape \color{commentcolor}]{--},

showstringspaces=false,

keepspaces=true,

morestring=[b]",
morestring=[d],

tabsize=3,

extendedchars=false,

sensitive=true,

breaklines=true,
breakatwhitespace=true,

basicstyle=\ttfamily\small,

captionpos=b,

columns=[l]fullflexible,

identifierstyle={\ttfamily\color{black}},

keywordstyle=[1]{\ttfamily\color{keywordcolor}},

keywordstyle=[2]{\ttfamily\color{sortcolor}},

keywordstyle=[3]{\ttfamily\color{tacticcolor}},

keywordstyle=[4]{\ttfamily\color{attributecolor}},

stringstyle=\ttfamily,

commentstyle={\ttfamily\footnotesize },

}

\endgroup

\tcbuselibrary{listings,breakable}

\newtcblisting{promptbox}{
  listing engine=listings,
  listing only, breakable,
  colback=gray!5, colframe=black!75, arc=3mm, boxrule=0.5pt,
  left=2mm, right=2mm, top=1mm, bottom=1mm,
  listing options={
    language=,                              
    basicstyle=\ttfamily\small\color{black},
    numbers=none, frame=none,              
    keywordstyle=\color{black},
    commentstyle=\color{black},
    stringstyle=\color{black},
    identifierstyle=\color{black},
    showstringspaces=false,
    breaklines=true,
    breakatwhitespace=false,
    breakindent=0pt,                       
    columns=fullflexible,
    keepspaces=true,
    xleftmargin=0pt, xrightmargin=0pt,      
  }
}

\newtcolorbox{promptbox2}{
  breakable,
  colback=gray!5,
  colframe=black!60,
  boxrule=0.6pt,
  arc=3mm,
  left=4mm,
  right=4mm,
  top=3mm,
  bottom=3mm
}

\newtcolorbox{rewritecase}[1]{
  breakable,
  colback=gray!3,
  colframe=black!35,
  colbacktitle=gray!14,
  coltitle=black,
  fonttitle=\bfseries\small,
  title={#1},
  boxrule=0.4pt,
  arc=1mm,
  left=2.5mm,
  right=2.5mm,
  top=1.5mm,
  bottom=1.5mm,
  before skip=5pt,
  after skip=5pt
}

\definecolor{sectioncolor}{HTML}{A91D3A}

\title{Lean Finder V2: Semantic Search for Mathlib That Disambiguates User Queries}

\author{
Jialin Lu\textsuperscript{1}\thanks{Equal contribution.}\quad
Xiaoyu Liu\textsuperscript{1}\footnotemark[1]\quad
Hengrui Luo\textsuperscript{2}\quad
Rodrigo Stehling\textsuperscript{1}\quad
Linxi Liu\textsuperscript{3}\\
\textbf{Weiran Sun\textsuperscript{1}\quad
Yuepeng Wang\textsuperscript{1}\quad
Wuyang Chen\textsuperscript{1}}\\
\textsuperscript{1}Simon Fraser University\quad
\textsuperscript{2}Rice University\quad
\textsuperscript{3}University of Pittsburgh
}

\newcommand{\std}[1]{{\scriptstyle\,\pm\,#1}}

\iclrfinalcopy 
\begin{document}

\maketitle

\begin{abstract}
We introduce \textbf{Lean Finder V2}, a semantic search framework for Mathlib that
disambiguates user queries through \emph{complementary model-level adversarial training and input-level query rewriting}.
Human mathematicians and coding agents rarely search Mathlib by providing detailed context. 
In practice, they query short properties, notation, aliases, or hypotheses, 
leaving much of the intended mathematics implicit.
In contrast, existing Mathlib search engines unrealistically assume context-rich queries.
The scarcity of real-world short queries paired with ground-truth Mathlib statements makes this ambiguity even more intractable.
We argue that this is essentially a \textbf{data} problem.
Through adversarial training, we scale up the generation of labeled short and ambiguous
queries, and learn to disambiguate user queries at both model-level fine-tuning and input-level query rewriting.
On the human-curated MathlibQR benchmark, our Lean Finder V2
sets a new state of the art, with at least 11\% relative improvement over the previous best Mathlib semantic search engines.
On top of our state-of-the-art retriever, optimized query rewriting further improves MathlibQR retrieval.
On unlabeled queries from human and coding-agent interactions, Lean Finder V2 is ranked as the strongest system.
Integrated with Claude Code, Lean Finder V2 improves proof success on downstream theorem-proving task. Lean Finder V2 is available at: \url{https://leanfinder.github.io}.
\end{abstract}


\section{Introduction}
\label{sec:introduction}

Advances in the Lean proof assistant and its central library Mathlib are turning
formal mathematics into a collaborative and verifiable workflow
\citep{de2015lean,moura2021lean}. Spanning a broad range of mathematics, Mathlib
now contains more than 400,000 theorems and definitions across roughly 10,000
files \citep{mathlibstats}. At this scale, effective theorem search is essential for both human and automated formalization, enabling users and systems to efficiently identify and reuse relevant results from Mathlib's large and growing library, 
and its quality shapes end-to-end proof success
\citep{lu2026lean,gao2026leansearchv2globalpremise,avigad2020mathematics}.
Alongside rapid progress in specialized Lean provers
\citep{ren2025deepseek,lin2025goedelproverv2scalingformaltheorem}, agentic systems have achieved tremendous success from competition mathematics to frontier research.
Many of these systems rely on Mathlib search engines to locate
reusable results during proof construction
\citep{liu2026numinaleanagentopengeneralagentic,
varambally2026hilbertrecursivelybuildingformal,
ju2026automatedconjectureresolutionformal, lean-lsp-mcp}.

Despite this progress, current theorem retrieval faces a central bottleneck:
\textbf{theorem-search queries from \emph{both} mathematicians and coding agents are inherently ambiguous}. 
Human mathematicians and coding agents rarely search by restating a complete theorem. 
They use short properties, notation, aliases, or hypotheses, and different users use different terms referring to the same mathematical result, leaving much
of the intended mathematics implicit. Such everyday queries are challenging precisely
because they expose wide \emph{ambiguity gaps}, yet existing systems leave these gaps unaddressed.
\citet{lu2026lean} synthesized realistic queries from intent clusters mined from Lean discussions, but the resulting data are static and generated independently of its retriever’s current failure modes. Other Lean search systems likewise rely largely on fixed, relatively detailed data, such as informalizations, formal statements, or proof states
\citep{gao2026leansearchv2globalpremise,gao2024semantic,asher2025leanexploresearchenginelean,yang2023leandojo}. We call such inputs \emph{context-rich}: they preserve much of the source
declaration's detail, unlike concise real-world queries that express the same
intent through a few salient cues.
Figure~\ref{fig:motivation}(a) quantifies this mismatch: human and coding-agent queries have median lengths of 16 and 15 tokens (i.e., ambiguous), respectively, compared with 82 tokens (i.e., context-rich) for the statement informalizations commonly
used for training previous Mathlib search engines. This gap is visible even when a shorter query
remains semantically aligned with the same target theorem:

\begin{tcolorbox}[
  title={\emph{One target, two levels of query ambiguity}},
  colback=gray!3,
  fontlower=\footnotesize\color{black!60},
  middle=0.8mm,
  colframe=black!45,
  colbacktitle=gray!14,
  coltitle=black,
  fonttitle=\bfseries\small,
  boxrule=0.5pt,
  arc=1mm,
  left=1.5mm,
  right=1.5mm,
  top=0.8mm,
  bottom=0.8mm,
  toptitle=0.6mm,
  bottomtitle=0.6mm]
\small
\textbf{Target statement:}
\begin{lstlisting}[
  language=lean,
  basicstyle=\ttfamily\footnotesize,
  numbers=none,
  frame=none,
  xleftmargin=0pt,
  framexleftmargin=0pt,
  aboveskip=2pt,
  belowskip=4pt]
theorem Function.Even.const
    {α β : Type*} [Neg α] (b : β) :
    Function.Even (fun _ : α ↦ b)
\end{lstlisting}
    \vspace{0.5em}
\textbf{Context-rich query:}
{\normalsize\textit{``for a type $\alpha$ with a negation and any type $\beta$, every \qA{constant
function} $f\colon\alpha\to\beta$, $f(x) = b$ for some $b \in \beta$, \qB{satisfies $f(-x) = f(x)$
for all $x \in \alpha$}, i.e., constant functions are even''}}\par
\vspace{1em}
\textbf{Ambiguous query:} {\normalsize\textit{``\qA{constant function} \qB{invariant under input negation}''}}
\tcblower
\textit{Note:} Same-colored spans express the same content in both queries. Context-rich query contains information that is not in ambiguous query.
\end{tcolorbox}

Both queries ask for the same mathematical fact, and the ambiguous query remains faithful to it.
Yet leaving the Mathlib concept of an even function implicit moves the target from rank one to rank
seven. Therefore, we ask our core question:

\begin{figure}[t]
  \centering
  \begin{minipage}[t]{0.47\linewidth}
    \centering
    \includegraphics[width=\linewidth]{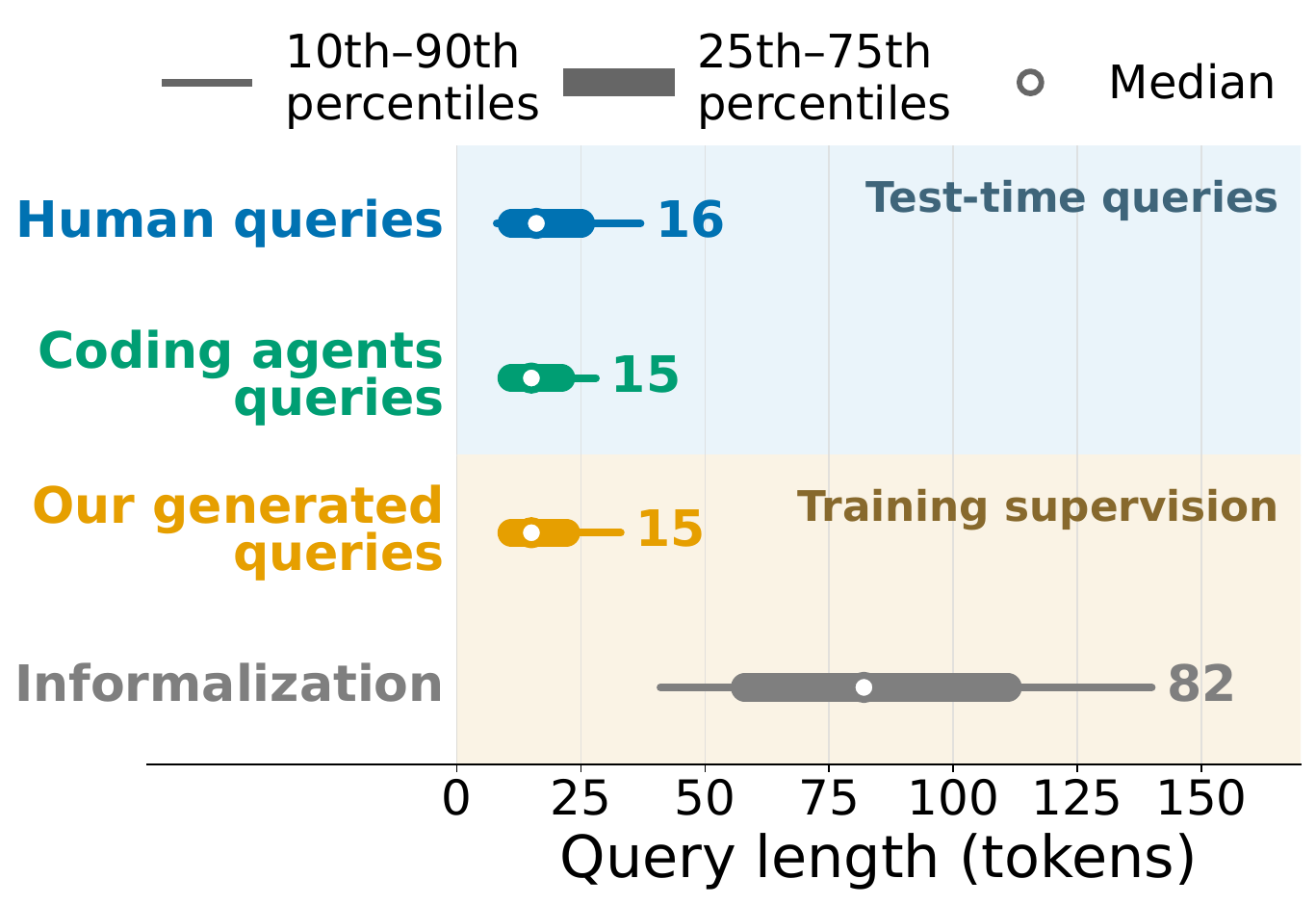}
    \small\textbf{(a)} Query-length distributions.
  \end{minipage}
  \hfill
  \begin{minipage}[t]{0.50\linewidth}
    \centering
    \includegraphics[width=\linewidth]{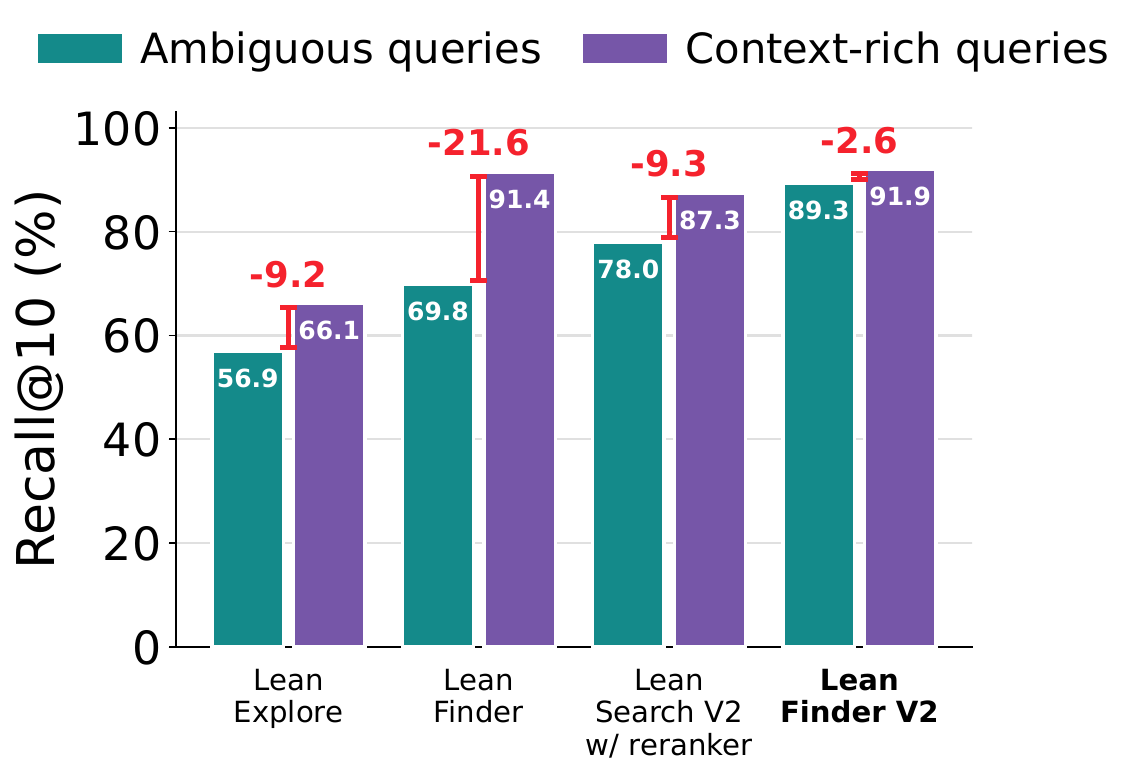}
    \small\textbf{(b)} Recall@10 across query types.
  \end{minipage}
  \caption{Real-world queries expose the ambiguity gap. (a) Human and
 coding-agent queries contain far less explicit context than statement informalizations which are commonly used to train Lean search engines, as shown by the length gap. Our generated queries used to train retrievers better match the length of real-world queries. Table~\ref{tab:query-length} shows more details.
  (b) Every existing Lean search engine performs worse on ambiguous queries compared to context-rich
ones, with Lean Finder V2 closing the gap to the smallest 2.6 percentage points and having the best performance on both query
types. This further demonstrates previous Lean search engines leave the ambiguity gap unaddressed.
  }
  \label{fig:motivation}
\vspace{-1em}
\end{figure}

\vspace{-0.5em}

\begin{center}
\fbox{
    \parbox{0.8\linewidth}{
        \centering
        \textit{How to disambiguate Mathlib queries, at both model level and input level?}
    }
}
\end{center}
\vspace{-0.5em}
In this paper, we introduce \textbf{Lean Finder V2}, a semantic search framework for Mathlib that excels
on ambiguous queries while retaining strong performance on context-rich queries. Lean Finder V2 addresses a \textbf{data bottleneck} caused by the \textbf{scarcity of labeled ambiguous queries}.
We synthesize ambiguous query-declaration pairs as training data to improve the retriever through \textbf{model-level adversarial training}, and the same synthetic data is used in \textbf{input-level prompt optimization} for query rewriting.
At the model level, a RL-trained attacker generates challenging and ambiguous queries for Mathlib declarations, with a \emph{difficulty reward} that targets weaknesses of the retriever and an \emph{alignment reward} that keeps each query faithful to its source declaration. These synthetic query-declaration pairs improve the retriever, whose feedback recalibrates query difficulty. At the input level, we use GEPA \citep{agrawal2026gepareflectivepromptevolution} to optimize a query-rewriting prompt on the same synthetic data with a ranking-based objective, making the intent of ambiguous queries explicit before retrieval. \textbf{Lean Finder V2 sets a new state of the art on MathlibQR} \citep{gao2026leansearchv2globalpremise}, with \textbf{14\% relative improvement} in Recall@10 over the previous best semantic search engine (Figure~\ref{fig:motivation}(b)). Lean Finder V2 ranks first among the evaluated Mathlib search engines and embedding models on the Legendre leaderboard for Mathlib retrieval.\footnote{\url{https://www.legendre-leaderboard.com/}, accessed October 22, 2026.} On top of the state-of-the-art retriever, the optimized rewriting prompt further improves retrieval on
MathlibQR. When integrated with Claude Code, Lean Finder V2 also improves downstream proof
success.

We summarize our contributions as follows:
\begin{enumerate}[leftmargin=*,itemsep=2pt,topsep=3pt]

  \item Lean Finder V2 sets a new state of the art on MathlibQR \citep{gao2026leansearchv2globalpremise} while retaining strong performance on context-rich queries \citep{lu2026lean}, achieving 14\% relative gains over the
previous-best semantic search system in Recall@10 on MathlibQR. On unlabeled queries for
Mathlib drawn from real user interactions, Lean Finder V2 is ranked as most helpful 18+\% more
often than the previous best system by frontier LLM judges (Claude Opus 5 and GPT-5.6 Sol).

  \item Through retriever-guided adversarial training, we construct 
  a training corpus of over 600K labeled ambiguous query-declaration pairs.
  The same corpus powers both model-level fine-tuning and input-level query-rewriting prompt optimization to disambiguate user queries. We will release this corpus to support future research.

  \item We present a study of
  input-side query rewriting.
  Using the same ambiguous query-declaration pairs for prompt optimization, GEPA-optimized query rewriting further improves MathlibQR retrieval for Lean Finder V2.

  \item Integrated with Claude
  Code, Lean Finder V2 improves downstream theorem proving, achieving a proof success rate of 39.3\% compared to the next-best 36.0\% on FATE-H \citep{jiang2026fateformalbenchmarkseries}.
\end{enumerate}

\section{Related Work}

Prior Lean retrieval systems improve access to Mathlib~\citep{gao2024semantic, gao2026leansearchv2globalpremise, asher2025leanexploresearchenginelean, lu2026lean, kurgan2026theoremgraphbridgingformalinformal}, and are often incorporated into agentic systems~\citep{varambally2026hilbertrecursivelybuildingformal, ju2026automatedconjectureresolutionformal, liu2026numinaleanagentopengeneralagentic}. Outside Lean, reinforcement learning (RL) has primarily optimized queries against fixed retrievers to improve retrieval accuracy~\citep{hsu2024groundingtryingllmsreinforcement, jiang2025deepretrieval}, rather than using generated queries as supervision to adapt the retriever itself. Lean Finder V2 instead adapts the retriever to real-world ambiguous queries through adversarial training: a generator produces declaration-faithful queries that challenge the current retriever, these query-declaration pairs retrain the retriever, and the updated retriever provides difficulty feedback for the next-round generator training. Complementarily, input-level query rewriting makes implicit mathematical context explicit before retrieval. Extended related work is provided in Appendix~\ref{app:related_work}.

\section{Methods}
\FloatBarrier

Lean Finder V2 disambiguates user queries by addressing a \emph{data} bottleneck: labeled ambiguous queries are scarce, and existing training data are context-rich and static (Section~\ref{sec:introduction}). We therefore generate ambiguous queries directly from Mathlib declarations through an adversarial training process that targets the current retriever's weaknesses, while ensuring each query remains faithful to its source.
Importantly, the generated ambiguous queries \emph{enable disambiguation at two complementary levels}.
At the \emph{model} level, we
train the retriever on the synthetic query-declaration pairs to recover the intended declaration directly from salient cues.
At the \emph{input} level, we use the same synthetic pairs to optimize a query-rewriting prompt that makes implicit intent explicit before retrieval.
Section~\ref{subsec:model-side-disambiguation} describes the adversarial training that generates the data and fine-tunes the retriever, and Section~\ref{subsec:input-side-disambiguation} reuses the data for input-level query rewriting.

\subsection{Model-level Disambiguation Through Adversarial Training}
\label{subsec:model-side-disambiguation}

The goal of model-level adversarial training is to synthesize high-quality ambiguous query-declaration pairs for retriever training. It couples a retrieval model with a query generator. The generator produces challenging and ambiguous queries that remain aligned with their source Mathlib declarations. We train the retriever on these query-declaration pairs, and its feedback guides the next round of query generator training.

\begin{figure}[t]
  \centering
  \includegraphics[width=\linewidth]{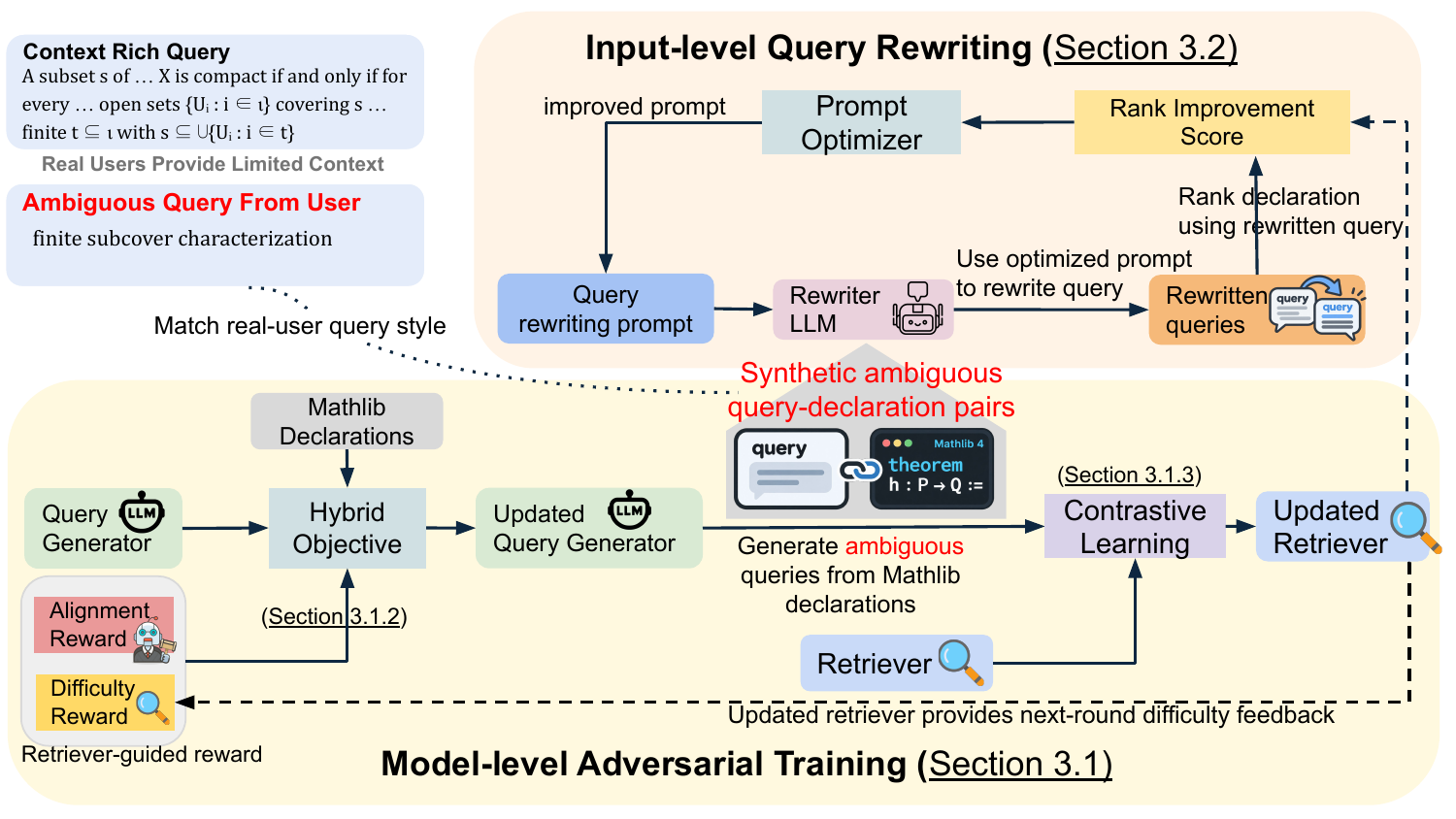}
  \caption{\textbf{Lean Finder V2 disambiguates Mathlib queries at two
  complementary levels.}
  \underline{Model-level adversarial training}
  (Section~\ref{subsec:model-side-disambiguation}) uses a hybrid objective that combines reinforcement learning and supervised fine-tuning with
 retriever-guided difficulty rewards to incentivize the query generator LLM to generate faithful, challenging, and ambiguous queries for Mathlib declarations. Contrastive learning on these query-declaration pairs improves the retriever, which supplies the query-difficulty feedback for the next training round of the generator.
  \underline{Input-level query rewriting}
  (Section~\ref{subsec:input-side-disambiguation}) uses the same generator-produced query-declaration pairs as the training data
  for optimizing the query-rewriting prompt, guided by the rank-improvement feedback based on the retriever. The optimized query rewriting prompt makes
  implicit mathematical context explicit before retrieval.}
  \label{fig:lean-finder-v2-pipeline}
\end{figure}

\emph{First}, we initialize the adversarial training process 
by training a retrieval model on large-scale, context-rich data, establishing broad Mathlib coverage before adapting it to ambiguous queries
(Section~\ref{subsec:base-retriever}). \emph{Second}, we optimize the generator with a retriever-guided reward that combines retrieval difficulty with query-declaration alignment, encouraging faithful queries that
challenge the current retriever (Section~\ref{subsec:query-generator}).
\emph{Finally}, we train the retriever on the generated query-declaration pairs and repeat the process with the updated retriever (Section~\ref{subsec:retriever-update}). Figure~\ref{fig:lean-finder-v2-pipeline} provides an overview of the framework, and Algorithm~\ref{alg:training-loop} summarizes the training procedure.

\subsubsection{Initializing and Fine-tuning the Retriever}
\label{subsec:base-retriever}
We initialize the retrieval model from Qwen3-Embedding-8B \citep{zhang2025qwen3embeddingadvancingtext}
and further fine-tune it on two sources of large-scale context-rich
data. First, we use synthetic user-query and Mathlib declaration pairs from~\citet{lu2026lean}, which capture multiple user intents summarized from Zulip discussions. Second, we curate training pairs from the informalization datasets from~\citet{lu2026lean} and LeanSearch V2 \citep{gao2026leansearchv2globalpremise}. Together, they form a large scale initial training set of query-declaration pairs.

We fine-tune the retriever using masked InfoNCE with in-batch negatives. To
reduce false-negative supervision, we mask any candidate whose cosine
similarity exceeds that of the labeled positive by more than a margin $m$.
This prevents highly plausible matches from being incorrectly treated as
negatives. The full objective is given in Appendix~\ref{app:retrieval_training_details}. This stage produces the initial retriever, which is used to guide the query-generator RL training (Section~\ref{subsec:query-generator}).

\subsubsection{Scaling Up the Generation of Ambiguous Queries via Adversarial Attacks}
\label{subsec:query-generator}

The initial retriever
from Section~\ref{subsec:base-retriever} provides the difficulty signal that guides the training of the query generator, whose role is to create ambiguous queries tailored to the retriever's weaknesses. These queries serve as ``adversarial'' samples for adapting the retriever to ambiguous inputs.
We use a hybrid
objective that combines on-policy reward optimization against the current
retriever with off-policy supervision from expert deployment queries, anchoring
generation in real-world query styles.

Given a Mathlib declaration $d$, the generator $\pi_\theta(\cdot\mid d)$, parameterized by $\theta$,
samples a group of candidate search queries $\{q_i\}_{i=1}^{G}$. For
the on-policy component, we use Group Relative Policy Optimization (GRPO)
\citep{shao2024deepseekmathpushinglimitsmathematical}. Each candidate $q_i$
receives a retriever-guided reward $r_i$ that combines its
alignment with $d$, difficulty for the current retriever, and length. We define
this reward in Equation~\ref{eq:query-reward}. The normalized advantage is:
\begin{equation}
  \widehat A_i =
  \frac{r_i-\operatorname{mean}(r_1,\ldots,r_G)}
       {\operatorname{std}(r_1,\ldots,r_G)+\epsilon_A}.
  \label{eq:grpo-advantage}
\end{equation}
We use the following GRPO loss:
{\small
\begin{align}
  \mathcal L_{\mathrm{GRPO}}(\theta)
  =-&\,\mathbb E\!\left[
  \frac{1}{G}\sum_{i=1}^{G}\frac{1}{|q_i|}
  \sum_{s=1}^{|q_i|}
  \left[
    \min\!\left(
      \rho_{i,s}(\theta)\widehat A_i,
      \operatorname{clip}\!\left(\rho_{i,s}(\theta),1-\epsilon,1+\epsilon\right)
      \widehat A_i
    \right)
    -\beta D_{\mathrm{KL}}^{i,s}
  \right]
  \right],
  \label{eq:grpo-loss}
\end{align}
}
where $s$ indexes token positions in query $q_i$,
$D_{\mathrm{KL}}^{i,s}$ is the per-token KL penalty to the reference
policy, and $\rho_{i,s}(\theta)$ is the token-level policy ratio,
\begin{equation}
  \rho_{i,s}(\theta)
  =\frac{
    \pi_\theta(q_{i,s}\mid d,q_{i,<s})
  }{
    \pi_{\theta_{\mathrm{old}}}(q_{i,s}\mid d,q_{i,<s})
  }.
  \label{eq:grpo-policy-ratio}
\end{equation}

\paragraph{Retriever-guided reward.}
For a generated query $q$
targeting Mathlib declaration $d$, let $k$ be the rank of
$d$ under the current retriever and let $L$ be the query length measured by the
retriever tokenizer. The reward is
\begin{equation}
  r(q,d)=
  \begin{cases}
    0, & \text{if the output format is invalid},\\
    \max\!\left\{0,\,
      j(q,d)f_{\mathrm{rank}}(k)
      -\lambda_{\mathrm{len}}p_{\mathrm{len}}(L)
    \right\}, & \text{otherwise},
  \end{cases}
  \label{eq:query-reward}
\end{equation}
where $j(q,d)\in\{0,1\}$ is an alignment gate, $f_{\mathrm{rank}}(k)$ measures query difficulty relative to the current retriever, and $p_{\mathrm{len}}(L)$ is a soft penalty for overlong queries, with $\lambda_{\mathrm{len}}$ controlling its strength.
The LLM judge based alignment gate prevents the generator from creating difficulty by producing a query
unrelated to the target declaration. Because the gate multiplies the difficulty reward, alignment is a
prerequisite rather than a property that can be traded against difficulty: an unrelated query receives
zero reward regardless of its retrieval rank. We define the rank-based difficulty reward below.
The difficulty term $f_{\mathrm{rank}}(k)$ selects queries in a useful rank region:
\begin{equation}
  f_{\mathrm{rank}}(k)=
  \begin{cases}
    0, & k\leq k_{\min}\ \text{or}\ k>k_{\mathrm{cut}},\\
    \dfrac{\log(k/k_{\min})}{\log(k_{\mathrm{sat}}/k_{\min})},
      & k_{\min}<k\leq k_{\mathrm{sat}},\\
    1, & k_{\mathrm{sat}}<k\leq k_{\mathrm{cut}}.
  \end{cases}
  \label{eq:rank-reward}
\end{equation}
Queries whose target appears within the top $k_{\min}$ receive no difficulty
reward because the retriever already ranks the target highly. The reward then
grows logarithmically and saturates across a useful challenging region. A
target outside the retrieval cutoff receives zero reward, preventing arbitrary
retrieval failure from being treated as useful difficulty. We set $k_{\min}=5$,
$k_{\mathrm{sat}}=30$, and $k_{\mathrm{cut}}=100$. Appendix~\ref{app:reward_hyperparameters} provides justifications for these
chosen thresholds. To prevent the generator from
increasing retrieval difficulty simply by appending unnecessary text, we apply a capped soft length penalty $p_{\mathrm{len}}(L)$ on the generated query if it becomes too long. Appendix~\ref{app:generator_training_details} provides the exact definition for $p_{\mathrm{len}}(L)$.

\paragraph{Hybrid objective with on-policy and supervised optimization.}
To anchor on-policy learning in real-world query styles, we curate the
supervised dataset from a corpus
of interactions by human users and coding agents with Lean search engines during formalization projects that rely on Mathlib. The diverse queries submitted during the interactions provide
initial supervision for generating queries that reflect how both human mathematicians and coding agents search
in practice. We use the standard SFT loss $\mathcal L_{\mathrm{SFT}}(\theta)$ on this dataset. Following the global
loss-mixing strategy of CHORD \citep{zhang2026onpolicyrlmeetsoffpolicy}, we
minimize $\mathcal L_{\mathrm{hybrid}}(\theta)=(1-\mu)
\mathcal L_{\mathrm{GRPO}}(\theta)+\mu\mathcal L_{\mathrm{SFT}}(\theta)$,
where $\mu$ decreases over
training, shifting the generator from supervised anchoring toward on-policy learning.

\subsubsection{Improving the Retriever With Generated Queries}
\label{subsec:retriever-update}

After training the generator against the current retriever, we generate queries from Mathlib declarations. We use the generated query-declaration pairs alongside the context-rich data from Section~\ref{subsec:base-retriever} to retrain the retriever with the same masked InfoNCE objective. The updated retriever then provides the rank signal for the next round of generator training. Algorithm~\ref{alg:training-loop} provides the pseudocode for the training process.

\subsection{Input-level Disambiguation Through Optimized Query Rewriting}
\label{subsec:input-side-disambiguation}

Complementary to model-level training in Section~\ref{subsec:model-side-disambiguation}, we can also disambiguate queries at the \emph{input} level (before retrieval) by using an LLM to rewrite them and expose their implicit mathematical context.
Specifically, using the retriever's ranking as feedback, we optimize the prompt given to the rewriting LLM.
This single optimized prompt is then reused to rewrite all queries.

\textbf{Importantly, it is precisely our synthesized ambiguous query-declaration pairs (generated in Section~\ref{subsec:query-generator}) that make this rewriting possible}.
Beyond retriever training (Section~\ref{subsec:retriever-update}), the \emph{same synthetic data} can also be \emph{reused} to discover a rewriting strategy (i.e., the prompt to the rewriting LLM) that exposes implicit mathematical context.
This is why we emphasize that the \ul{same data bottleneck connects both model-level and input-level disambiguation}.

We use GEPA \citep{agrawal2026gepareflectivepromptevolution} to optimize a rewriting prompt initialized from a simple seed prompt.
We split a subset of the synthetic query-declaration pairs from Section~\ref{subsec:query-generator} into feedback and Pareto sets.  In each optimization step, GEPA selects a candidate prompt to rewrite a minibatch of queries from the feedback set, and the fixed Lean Finder V2 retriever ranks each target declaration. The resulting ranks yield a scalar score $s$ (defined in Equation~\ref{eq:gepa-retrieval-score}) and textual feedback comparing the target ranks for the original query, seed-prompt rewrite, and current rewrite (feedback template is in Appendix~\ref{prompt:gepa_feedback}). A reflection model (DeepSeek-V4.1-Flash~\citep{deepseekai2026deepseekv41flashpushinglimitskv}) uses this feedback to revise the prompt.
GEPA retains a revision only if it improves the mean score on the same minibatch, then evaluates it on every Pareto-set example and adds it to the candidate prompt pool. When selecting the next prompt to revise from the pool, GEPA favors candidates that perform best on more Pareto-set examples. After optimization, it returns the prompt with the highest average Pareto-set score.

The score $s$ rewards a rewrite both for placing the target declaration near the top of the retrieved list and for placing it at a better rank than the original query and the seed-prompt rewrite do. For each query-declaration pair $(q_i,d_i)$ and rewriting prompt $p$, let
$rank_i(p)$ denote the rank of the target declaration $d_i$ after rewriting
$q_i$ with $p$; $rank_i^{\mathrm{orig}}$ and $rank_i^{\mathrm{seed}}$ denote
its ranks for the original query and seed-prompt rewrite. The per-example
scalar score $s_i(p)$ is
{\footnotesize
\begin{equation}
\label{eq:gepa-retrieval-score}
s_i(p)
=
\frac{\lambda_{\mathrm{abs}}}{\log_2(rank_i(p)+1)}
+\lambda_{\mathrm{orig}}
\frac{rank_i^{\mathrm{orig}}}
{rank_i^{\mathrm{orig}}+rank_i(p)}
+\lambda_{\mathrm{seed}}
\frac{rank_i^{\mathrm{seed}}}
{rank_i^{\mathrm{seed}}+rank_i(p)}.
\end{equation}
}
The three terms reward absolute retrieval quality, improvement over the original query, and improvement over the seed-prompt rewrite, respectively, and $\lambda_{\mathrm{abs}}, \lambda_{\mathrm{orig}}, \lambda_{\mathrm{seed}}$ are weights that balance them. See Appendix~\ref{app:prompt_opt_details} for details on how we use GEPA. The optimized prompt is shown in Appendix~\ref{app:gepa-prompt}.

\section{Experiments}
\label{sec:experiments}
We initialize the retrieval model from Qwen3-Embedding-8B
\citep{zhang2025qwen3embeddingadvancingtext} and the query generator from
Qwen3-4B \citep{yang2025qwen3technicalreport}, with GPT-OSS-120B
\citep{openai2025gptoss120bgptoss20bmodel} as the alignment judge, and
DeepSeek-V4.1-Flash \citep{deepseekai2026deepseekv41flashpushinglimitskv} for both GEPA prompt optimization and query rewriting. We perform two rounds of alternating generator-retriever training. In each round, we optimize the query generator on $2{,}000$ Mathlib statements across diverse modules to encourage broad coverage of mathematical domains while keeping training tractable. Full model training and prompt optimization details are in Appendix~\ref{app:training_details}, and prompts are in Appendix~\ref{app:prompts}.

We evaluate Lean Finder V2 on
both concise and context-rich query benchmarks, comparing against existing Lean search engines (Section~\ref{sec:retrieval-results}). We test retrieval quality on real-world unlabeled queries collected from human and coding-agent interactions with Lean search engines (Section~\ref{sec:deployment-preference}). We then evaluate input-side disambiguation by comparing our GEPA-optimized query rewriting against existing query-rewriting methods from Lean search engines on concise queries (Section~\ref{sec:gepa-results}). We assess downstream application by comparing Claude Code equipped with Lean Finder V2 against the same agent with no semantic search tools or a prior retrieval system (Section~\ref{sec:proving-results}). Finally, Appendix~\ref{sec:ablations} reports ablation studies that isolate the contributions  of the training signals.

\subsection{Lean Finder V2 Is the Strongest Retriever Across Query Types}
\label{sec:retrieval-results}

\paragraph{Setup.}
We evaluate concise and ambiguous query retrieval on the human-curated MathlibQR benchmark
(the 810-query fair set) \citep{gao2026leansearchv2globalpremise}, and
context-rich query retrieval on the 2{,}526-query test set of \citet{lu2026lean}.
We ensure that the ground-truth declaration for every evaluation query is present in the corpus of every evaluated Lean search engine.
\textbf{We ensure that neither any ground-truth declaration from the test sets nor any query associated with such a declaration exists in the training corpus at any stage of training.}

We report three checkpoints in our series: \emph{Lean Finder V2 Round 0}, trained only on
existing context-rich data (Section~\ref{subsec:base-retriever});
\emph{Lean Finder V2 Round 1}, which adds queries from the first RL-trained generator (Section~\ref{subsec:query-generator}); and \emph{Lean Finder V2}, the final model after two rounds. We compare them with Lean Finder \citep{lu2026lean}, LeanSearch V2 \citep{gao2026leansearchv2globalpremise}, LeanExplore \citep{asher2025leanexploresearchenginelean}, TheoremGraph \citep{kurgan2026theoremgraphbridgingformalinformal}, and Claude Code (Sonnet 5) \citep{anthropic2026claudesonnet5}. For TheoremGraph, we report the MathlibQR results from \citet{kurgan2026theoremgraphbridgingformalinformal}; since its evaluation uses a cutoff of 10, Recall@50 and Recall@100 are not reported. We instruct Claude Code to return only ten declarations per query due to cost, so Recall@50 and Recall@100 are also not reported.
Appendix~\ref{app:retrieval_evaluation_setup} provides the full protocol.

\paragraph{Metrics.}
Following prior evaluations of Lean search engines
\citep{lu2026lean,gao2026leansearchv2globalpremise,asher2025leanexploresearchenginelean}, we report nDCG@$k$ and Recall@$k$. nDCG@$k$ rewards placing the target declaration near
the top of the first $k$ results, while Recall@$k$ measures the proportion of queries for which the target appears within those results.

\begin{table}[h!]
\centering
\captionsetup{font=small}
\caption{\textbf{Retrieval performance on the 810-query MathlibQR fair set of short, human-written queries.} Systems in comparison include LeanExplore, LeanSearch V2, Claude Code (Sonnet 5), TheoremGraph, Lean Finder, and three Lean Finder V2 variants (Round 0,
Round 1, and the final model). Lean Finder V2 denotes the final model. As TheoremGraph uses cutoff of 10 in~\citet{kurgan2026theoremgraphbridgingformalinformal}, its Recall@50 and Recall@100 are not reported. Claude Code returns
only ten candidates, so its Recall@50 and Recall@100 are also not reported. We also report the model
size of each system in comparison. Best results in each metric column are \textbf{bold}.}
\label{tab:mathlibqr-retrieval}
\vspace{-0.5em}
{\small
\renewcommand{\arraystretch}{0.95}
\begin{tabular*}{\linewidth}{@{\extracolsep{\fill}}lcccc@{\hspace{1.25em}}ccc@{}}
\toprule
 & & \multicolumn{3}{c}{nDCG@$k$} & \multicolumn{3}{c}{Recall@$k$} \\
\cmidrule(lr){3-5} \cmidrule(lr){6-8}
System & Model size & 1 & 5 & 10 & 10 & 50 & 100 \\
\midrule
LeanExplore                         & 109M & 0.246 & 0.358 & 0.393 & 0.569 & 0.743 & 0.789 \\
LeanSearch V2 (retriever-only)      & 8B & 0.340 & 0.472 & 0.494 & 0.657 & 0.790 & 0.830 \\
LeanSearch V2 (retriever+reranker)    & 8B + 8B & 0.470 & 0.601 & 0.623 & 0.780 & 0.847 & 0.858 \\
Claude Code (Sonnet 5)              & -- & 0.527 & 0.629 & 0.641 & 0.747 & -- & -- \\
TheoremGraph & 8B & 0.391 & 0.534 & 0.558 & 0.733 & -- & -- \\
Lean Finder                         & 7B & 0.370 & 0.514 & 0.533 & 0.698 & 0.824 & 0.875 \\
\midrule
Lean Finder V2 Round 0                  & 8B & 0.463 & 0.590 & 0.611 & 0.760 & 0.864 & 0.901 \\
Lean Finder V2 Round 1                  & 8B & 0.520 & 0.657 & 0.682 & 0.842 & 0.925 & 0.951 \\
\textbf{Lean Finder V2}                 & 8B & \textbf{0.558} & \textbf{0.707} & \textbf{0.732} & \textbf{0.893} & \textbf{0.956} & \textbf{0.973} \\
\bottomrule
\end{tabular*}
}
\end{table}

\begin{table}[h!]
\centering
\captionsetup{font=small}
\caption{\textbf{Retrieval performance on the 2{,}526-query context-rich test set.} Systems in comparison include LeanExplore, LeanSearch V2, Claude Code (Sonnet 5), Lean Finder, and three Lean Finder V2 variants (Round 0, Round 1, and the final model). Lean Finder V2 denotes the final model. Claude Code returns only ten candidates, so its Recall@50 and Recall@100 are not reported. Best results in each column are \textbf{bold}.}
\label{tab:information-rich-retrieval}
\vspace{-0.5em}
{\small
\renewcommand{\arraystretch}{0.95}
\begin{tabular*}{0.9\linewidth}{@{\extracolsep{\fill}}lccc@{\hspace{1.25em}}ccc@{}}
\toprule
 & \multicolumn{3}{c}{nDCG@$k$} & \multicolumn{3}{c}{Recall@$k$} \\
\cmidrule(lr){2-4} \cmidrule(lr){5-7}
System & 1 & 5 & 10 & 10 & 50 & 100 \\
\midrule
LeanExplore                         & 0.180 & 0.376 & 0.406 & 0.661 & 0.837 & 0.872\\
LeanSearch V2 (retriever-only)      & 0.414 & 0.569 & 0.596 & 0.783 & 0.907 & 0.942 \\
LeanSearch V2 (retriever+reranker)    & 0.506 & 0.669 & 0.691 & 0.873 & 0.954 & 0.964 \\
Claude Code (Sonnet 5)    & 0.245 & 0.345 & 0.363 & 0.485 & -- & -- \\
Lean Finder                      & 0.544 & 0.720 & 0.742 & 0.914 & 0.980 & 0.989 \\
\midrule
Lean Finder V2 Round 0                  & 0.563 & 0.720 & 0.743 & 0.910 & 0.967 & 0.981 \\
Lean Finder V2 Round 1                  & 0.569 & \textbf{0.728} & 0.743 & 0.916 & \textbf{0.985} & \textbf{0.993} \\
\textbf{Lean Finder V2}                 & \textbf{0.573} & \textbf{0.728} & \textbf{0.749} & \textbf{0.919} & \textbf{0.985} & 0.992 \\
\bottomrule
\end{tabular*}
}
\end{table}

Tables~\ref{tab:mathlibqr-retrieval} and
\ref{tab:information-rich-retrieval} show that Lean Finder V2, which is a retriever-only system, achieves the best result in every reported metric on MathlibQR, and performs competitively on context-rich query benchmark, outperforming LeanSearch V2, previous best two-stage retriever+reranker Lean semantic search engine. We discuss model sizes in Appendix~\ref{app:model_scale}. We attribute Claude Code's strong MathlibQR performance to Sonnet 5 combining pretrained knowledge of Mathlib naming with iterative search over Mathlib documentation to infer Mathlib terminology from ambiguous and concise queries. Context-rich queries instead require fine-grained matching of a detailed mathematical description against many closely related Mathlib declarations, which directly favors dense retrievers trained on paired informalization-declaration data over general-purpose agentic search.

Lean Finder V2 Round 0, trained only on context-rich data, performs strongly on context-rich queries and competitively on ambiguous queries, but performs worse than LeanSearch V2 with reranking on MathlibQR. Adversarial training closes this gap by producing large-scale synthetic ambiguous queries: training on these pairs improves every MathlibQR metric from Round 0 to the final model, while context-rich performance remains consistently strong, demonstrating the quality and usefulness of our synthetic data. Appendix~\ref{sec:ablations} reports ablation studies that further isolate the
contributions of this training procedure. Finally, Appendix~\ref{app:reraker_results} shows that reranking further improves retrieval performance on MathlibQR and context-rich queries on top of the strong Lean Finder V2 performance.

\subsection{Lean Finder V2 Is Preferred on Unlabeled Real-World Queries}
\label{sec:deployment-preference}

Ground-truth declarations are unavailable for most queries in practice. We therefore complement labeled benchmarks with preference evaluation on 200
queries collected from human and coding-agent interactions with Lean search engines. These evaluation queries are drawn from interactions held out from the generator's SFT data in Section~\ref{subsec:query-generator}. 
Each retrieval system retrieves five declarations per query. Following prior work~\citep{lu2026lean, gao2026leansearchv2globalpremise, asher2025leanexploresearchenginelean}, we use LLM judges to determine the quality and helpfulness of retrieval from different Lean search systems. Two frontier models, GPT-5.6 Sol and Claude Opus
5, independently rank the five systems by the helpfulness of their retrieved declarations to the query. We ensure that the identity of each system is not visible to LLM judges, and the order of systems is shuffled per query to avoid positional bias. Appendix~\ref{app:llm-judge-validation} gives the evaluation details and validates the reliability of both LLM judges against the human-curated MathlibQR
labels.

\begin{table}[h!]
\centering
\captionsetup{font=small}
\caption{\textbf{LLM-judged retrieval preference on 200 unlabeled deployment queries.}
GPT-5.6 Sol and Claude Opus 5 independently rank LeanExplore, LeanSearch V2,
Claude Code (Sonnet 5), Lean Finder, and Lean Finder V2. Rank
1 is reported as count (percentage); lower mean rank is better. Best results
are \textbf{bold}.}
\label{tab:deployment-preference}
\vspace{-1em}
{\small
\renewcommand{\arraystretch}{0.95}
\setlength{\tabcolsep}{5pt}
\begin{tabular}{@{}lcc@{\hspace{1.2em}}cc@{}}
\toprule
& \multicolumn{2}{c}{GPT-5.6 Sol} & \multicolumn{2}{c}{Claude Opus 5} \\
\cmidrule(lr){2-3}\cmidrule(lr){4-5}
System & Mean rank $\downarrow$ & Rank 1 & Mean rank $\downarrow$ & Rank 1 \\
\midrule
LeanExplore & 3.740 & 14 (7.0\%) & 3.735 & 8 (4.0\%) \\
LeanSearch V2 (retriever+reranker) & 2.380 & 52 (26.0\%) & 2.305 & 64 (32.0\%) \\
Claude Code (Sonnet 5) & 3.775 & 32 (16.0\%) & 3.920 & 22 (11.0\%) \\
Lean Finder & 3.125 & 32 (16.0\%) & 3.095 & 30 (15.0\%) \\
\textbf{Lean Finder V2} & \textbf{1.980} & \textbf{70 (35.0\%)} &
  \textbf{1.945} & \textbf{76 (38.0\%)} \\
\bottomrule
\end{tabular}
}
\vspace{-1em}
\end{table}

Table~\ref{tab:deployment-preference} shows that both judges place Lean Finder V2 first overall: its mean rank is
$1.980$ under GPT-5.6 Sol and $1.945$ under Claude Opus 5, and it receives the
most rank-1 selections from each judge. The two judges choose the same
rank-1 system on 89.0\% of queries. Thus, the gain on labeled benchmarks
extends to the open-ended queries that humans and coding agents issue during
deployment.

\subsection{Optimized Query Rewriting Improves Top-Ranked Retrieval}
\label{sec:gepa-results}

On MathlibQR, we compare GEPA-optimized rewriting \citep{agrawal2026gepareflectivepromptevolution} against its seed prompt, no rewriting, LeanSearch\footnote{\url{https://leansearch.net/}} and LeanDex\footnote{\url{https://leandex.projectnumina.ai/}} augmentations. All configurations use the same Lean Finder V2 retriever. For each query-rewriting method, we independently generate three rewrites per query and report mean and standard deviation in all metrics; the no-rewriting configuration is evaluated once, so no standard deviation is reported. We report the same retrieval metrics from Section~\ref{sec:retrieval-results}. See Appendix~\ref{app:query_rewriting_evaluation_setup} for the full protocol.

\begin{table}[htbp]
\centering
\captionsetup{font=small}
\caption{\textbf{Query-rewriting performance on the 810-query MathlibQR fair set}. We compare GEPA-optimized rewriting with no rewriting, LeanSearch and LeanDex augmentation, and the GEPA seed prompt rewriting. Each rewriting method produces three query sets, with the Lean Finder V2 retriever fixed; results report their mean and standard deviation, while the no-rewriting condition is evaluated once. Best results in each column are \textbf{bold}.}
\label{tab:gepa-rewriting}
\vspace{-0.5em}

\setlength{\tabcolsep}{4pt}
\renewcommand{\arraystretch}{1.1}
\resizebox{\linewidth}{!}{%
\begin{tabular}{@{}lcccccc@{}}
\toprule
& \multicolumn{3}{c}{nDCG@$k$}
& \multicolumn{3}{c}{Recall@$k$} \\
\cmidrule(lr){2-4}\cmidrule(lr){5-7}
Input Rewrite Config.
& 1 & 5 & 10
& 10 & 50 & 100 \\
\midrule
Original query (no rewrite)
& $0.558$ & $0.707$ & $0.732$
& $0.893$ & $0.956$ & $0.973$ \\

LeanSearch augmentation
& $0.475\std{0.006}$ & $0.612\std{0.004}$ & $0.637\std{0.004}$
& $0.808\std{0.004}$ & $0.908\std{0.005}$ & $0.936\std{0.005}$ \\

LeanDex augmentation
& $0.520\std{0.013}$ & $0.665\std{0.009}$ & $0.688\std{0.009}$
& $0.852\std{0.008}$ & $0.944\std{0.005}$ & $0.962\std{0.005}$ \\


GEPA seed rewriting
& $0.532\std{0.002}$ & $0.674\std{0.004}$ & $0.695\std{0.005}$
& $0.854\std{0.010}$ & $0.942\std{0.004}$ & $0.963\std{0.004}$ \\

\textbf{GEPA-optimized rewriting}
& $\mathbf{0.588}\std{0.004}$
& $\mathbf{0.722}\std{0.003}$
& $\mathbf{0.741}\std{0.002}$
& $\mathbf{0.895}\std{0.002}$
& $\mathbf{0.964}\std{0.007}$
& $\mathbf{0.975}\std{0.003}$ \\
\bottomrule
\end{tabular}%
}
\end{table}

Table~\ref{tab:gepa-rewriting} shows that GEPA-optimized rewriting achieves the strongest top-ranked retrieval, raising nDCG@1 from 0.558 to 0.588 over the original queries. With Recall@100 already at 0.973, rewriting mainly resolves which retrieved candidate the user intends, moving the ground-truth declaration into the top ranks that users inspect first. The improvement that the GEPA-optimized rewriting brings further demonstrates the value of our synthetic ambiguous query-declaration pairs for addressing real-world queries from a data perspective: the same synthetic data is used for both training the retrieval model and for optimizing a rewriting prompt, and each improves retrieval on human-written MathlibQR queries. LeanSearch, LeanDex, and the seed prompt reduce performance at every cutoff. We find that their rewrites tend to add unnecessary context and shift focus away from the user's intent. We present a case study in Appendix~\ref{app:rewrite_example}. In addition, with reranking added on top of query rewriting, we achieve further improvement, reaching 0.610 for nDCG@1. Appendix~\ref{app:reraker_results} provides more details for the reranking+rewriting configuration.
\subsection{Lean Finder V2 Helps the Most on Theorem Proving Task}
\label{sec:proving-results}

\begin{wraptable}{r}{0.5\linewidth}
\vspace{-0.8\baselineskip}
\centering
\captionsetup{font=small}
\caption{\textbf{Theorem-proving performance on FATE-H
benchmark.} Mean pass rates and standard deviations of Claude Code (Sonnet 5) over three runs under matched agent settings and
budgets; best result is \textbf{bold}.}
\label{tab:fate-h-proving}
{\small
\renewcommand{\arraystretch}{0.95}
\begin{tabular*}{\linewidth}{@{\extracolsep{\fill}}lc@{}}
\toprule
Retrieval tool & Pass rate (\%) \\
\midrule
No semantic retrieval tool & $34.0\std{1.0}$ \\
LeanExplore & $27.3\std{0.6}$ \\
Lean Finder & $30.7\std{1.2}$ \\
LeanSearch V2 (retriever+reranker) & $36.0\std{1.0}$ \\
\textbf{Lean Finder V2} & $\mathbf{39.3}\std{1.2}$ \\
\bottomrule
\end{tabular*}
}
\vspace{-0.8\baselineskip}
\end{wraptable}

We evaluate Claude Code (Sonnet 5) on FATE-H \citep{jiang2026fateformalbenchmarkseries}, which contains difficult formal algebra problems, using LeanExplore, LeanSearch V2 with reranking, Lean Finder, Lean Finder V2, and no additional semantic search tool. All settings share the same prompt template and per-problem budget. Table~\ref{tab:fate-h-proving} reports mean pass rates and standard deviations over three runs, counting only compiling, sorry-free proofs; Appendix~\ref{app:theorem_proving_evaluation_setup} details the protocol.

Lean Finder V2 achieves the highest pass rate at $39.3\%$. The no semantic search tool baseline is also strong and outperforms LeanExplore and Lean Finder while relying on Sonnet 5's internal Mathlib knowledge and repository grep. This suggests that, under a fixed agent budget, retrieval improves theorem proving only when it returns sufficiently relevant results to justify the additional effort required to inspect them and invoke the tool. Only LeanSearch V2 and Lean Finder V2 outperform Claude Code with no semantic search tool. The result shows that Lean Finder V2's training strategy is consistent with queries issued during agentic proof development and translates its retrieval advantage into stronger end-to-end theorem proving.

\section{Conclusion}
We introduce Lean Finder V2, a semantic search framework designed around how human mathematicians and coding agents actually search Mathlib: with ambiguous queries that leave much of their intent implicit. Lean Finder V2 aligns the retriever with real-world queries through complementary model-level adversarial training and input-level query-rewriting prompt optimization. The same synthetic ambiguous query-declaration pairs power both optimization processes. Comprehensive evaluations demonstrate its superior performance over existing search systems. By making Mathlib more accessible through the concise and ambiguous queries used in practice, Lean Finder V2 provides critical infrastructure for autonomous agents in formal mathematics and scalable human-AI collaboration.

\section*{Acknowledgment}

This research used resources of the National Energy Research Scientific Computing Center, a DOE Office of Science User Facility supported by the Office of Science of the U.S. Department of Energy under Contract No. DE-AC02-05CH11231 using NERSC award NERSC DDR-ERCAP0034682 and ASCR-ERCAP0031463, and U.S. National Science Foundation NSF Grant DMS-2606336, DMS-2616828.


\FloatBarrier

\bibliography{iclr2027_conference}

\begin{thebibliography}{34}
\providecommand{\natexlab}[1]{#1}
\providecommand{\url}[1]{\texttt{#1}}
\expandafter\ifx\csname urlstyle\endcsname\relax
  \providecommand{\doi}[1]{doi: #1}\else
  \providecommand{\doi}{doi: \begingroup \urlstyle{rm}\Url}\fi

\bibitem[Agrawal et~al.(2026)Agrawal, Tan, Soylu, Ziems, Khare, Opsahl-Ong, Singhvi, Shandilya, Ryan, Jiang, Potts, Sen, Dimakis, Stoica, Klein, Zaharia, and Khattab]{agrawal2026gepareflectivepromptevolution}
Lakshya~A Agrawal, Shangyin Tan, Dilara Soylu, Noah Ziems, Rishi Khare, Krista Opsahl-Ong, Arnav Singhvi, Herumb Shandilya, Michael~J Ryan, Meng Jiang, Christopher Potts, Koushik Sen, Alexandros~G. Dimakis, Ion Stoica, Dan Klein, Matei Zaharia, and Omar Khattab.
\newblock Gepa: Reflective prompt evolution can outperform reinforcement learning, 2026.
\newblock URL \url{https://arxiv.org/abs/2507.19457}.

\bibitem[{Anthropic}(2026)]{anthropic2026claudesonnet5}
{Anthropic}.
\newblock Claude sonnet 5 system card, June 2026.
\newblock URL \url{https://www-cdn.anthropic.com/9e6a1044980d8c4ed85669faf9c2a8342e2e9f1e/Claude%20Sonnet%205%20System%20Card.pdf}.

\bibitem[Asher(2025)]{asher2025leanexploresearchenginelean}
Justin Asher.
\newblock Leanexplore: A search engine for lean 4 declarations, 2025.
\newblock URL \url{https://arxiv.org/abs/2506.11085}.

\bibitem[Avigad \& Massot(2020)Avigad and Massot]{avigad2020mathematics}
Jeremy Avigad and Patrick Massot.
\newblock Mathematics in lean, 2020.

\bibitem[Cao et~al.(2026)Cao, Qiu, Liu, Wang, Guo, Feng, Zhi, and Gao]{cao2026mechmathagentteamllm}
Yichuan Cao, Ruichen Qiu, Junqi Liu, Jiaqi Wang, Dakai Guo, Ruyong Feng, Lihong Zhi, and Xiao-Shan Gao.
\newblock Mechmath agent team: Llm driven agents for mathematical research, 2026.
\newblock URL \url{https://arxiv.org/abs/2607.04394}.

\bibitem[De~Moura et~al.(2015)De~Moura, Kong, Avigad, Van~Doorn, and von Raumer]{de2015lean}
Leonardo De~Moura, Soonho Kong, Jeremy Avigad, Floris Van~Doorn, and Jakob von Raumer.
\newblock The lean theorem prover (system description).
\newblock In \emph{Automated Deduction-CADE-25: 25th International Conference on Automated Deduction, Berlin, Germany, August 1-7, 2015, Proceedings 25}, pp.\  378--388. Springer, 2015.

\bibitem[{DeepSeek-AI}(2026)]{deepseekai2026deepseekv41flashpushinglimitskv}
{DeepSeek-AI}.
\newblock Deepseek-v4.1-flash: Pushing the limits of kv cache compression, 2026.
\newblock URL \url{https://arxiv.org/abs/2609.19969}.

\bibitem[Dressler(2025)]{lean-lsp-mcp}
Oliver Dressler.
\newblock {Lean LSP MCP: Tools for agentic interaction with the Lean theorem prover}, 3 2025.
\newblock URL \url{https://github.com/oOo0oOo/lean-lsp-mcp}.

\bibitem[Gao et~al.(2024)Gao, Ju, Jiang, Qin, and Dong]{gao2024semantic}
Guoxiong Gao, Haocheng Ju, Jiedong Jiang, Zihan Qin, and Bin Dong.
\newblock A semantic search engine for mathlib4.
\newblock \emph{arXiv preprint arXiv:2403.13310}, 2024.

\bibitem[Gao et~al.(2026)Gao, Sun, Jiang, Wang, Xu, Wu, Dai, and Dong]{gao2026leansearchv2globalpremise}
Guoxiong Gao, Zeming Sun, Jiedong Jiang, Yutong Wang, Jingda Xu, Peihao Wu, Bryan Dai, and Bin Dong.
\newblock Leansearch v2: Global premise retrieval for lean 4 theorem proving, 2026.
\newblock URL \url{https://arxiv.org/abs/2605.13137}.

\bibitem[Hsu et~al.(2024)Hsu, Khattab, Finn, and Sharma]{hsu2024groundingtryingllmsreinforcement}
Sheryl Hsu, Omar Khattab, Chelsea Finn, and Archit Sharma.
\newblock Grounding by trying: Llms with reinforcement learning-enhanced retrieval, 2024.
\newblock URL \url{https://arxiv.org/abs/2410.23214}.

\bibitem[Hu et~al.(2021)Hu, Shen, Wallis, Allen-Zhu, Li, Wang, Wang, and Chen]{lora}
Edward~J. Hu, Yelong Shen, Phillip Wallis, Zeyuan Allen-Zhu, Yuanzhi Li, Shean Wang, Lu~Wang, and Weizhu Chen.
\newblock Lora: Low-rank adaptation of large language models, 2021.
\newblock URL \url{https://arxiv.org/abs/2106.09685}.

\bibitem[Jagatap et~al.(2025)Jagatap, Anand, Singh, and Comar]{jagatap2025reinforcement}
Akshay Jagatap, Neeraj Anand, Sonali Singh, and Prakash~Mandayam Comar.
\newblock Reinforcement learning for adversarial query generation to enhance relevance in cold-start product search.
\newblock In \emph{Proceedings of the 63rd Annual Meeting of the Association for Computational Linguistics (Volume 6: Industry Track)}, pp.\  1300--1307, 2025.

\bibitem[Jiang et~al.(2026)Jiang, He, Wang, Gao, Hu, Wang, Guan, Wu, Dai, Xiao, and Dong]{jiang2026fateformalbenchmarkseries}
Jiedong Jiang, Wanyi He, Yuefeng Wang, Guoxiong Gao, Yongle Hu, Jingting Wang, Nailin Guan, Peihao Wu, Chunbo Dai, Liang Xiao, and Bin Dong.
\newblock Fate: A formal benchmark series for frontier algebra of multiple difficulty levels, 2026.
\newblock URL \url{https://arxiv.org/abs/2511.02872}.

\bibitem[Jiang et~al.(2025)Jiang, Lin, Cao, Tian, Kang, Wang, Sun, and Han]{jiang2025deepretrieval}
Pengcheng Jiang, Jiacheng Lin, Lang Cao, Runchu Tian, SeongKu Kang, Zifeng Wang, Jimeng Sun, and Jiawei Han.
\newblock Deepretrieval: Hacking real search engines and retrievers with large language models via reinforcement learning.
\newblock \emph{arXiv preprint arXiv:2503.00223}, 2025.

\bibitem[Ju et~al.(2026)Ju, Gao, Jiang, Wu, Sun, Liu, Chen, Wang, Wang, Wang, He, Wu, Xiao, Liu, Dai, and Dong]{ju2026automatedconjectureresolutionformal}
Haocheng Ju, Guoxiong Gao, Jiedong Jiang, Bin Wu, Zeming Sun, Shurui Liu, Leheng Chen, Yutong Wang, Yuefeng Wang, Zichen Wang, Wanyi He, Peihao Wu, Liang Xiao, Ruochuan Liu, Bryan Dai, and Bin Dong.
\newblock Automated conjecture resolution with formal verification, 2026.
\newblock URL \url{https://arxiv.org/abs/2604.03789}.

\bibitem[Kurgan et~al.(2026)Kurgan, Wang, Leonen, Szeto, Alexander, Remizov, Alper, Inchiostro, and Ilin]{kurgan2026theoremgraphbridgingformalinformal}
Simon Kurgan, Evan Wang, Eric Leonen, Sophie Szeto, Luke Alexander, Artemii Remizov, Jarod Alper, Giovanni Inchiostro, and Vasily Ilin.
\newblock Theoremgraph: Bridging formal and informal mathematics, 2026.
\newblock URL \url{https://arxiv.org/abs/2606.25363}.

\bibitem[{Lean Prover Community}(2026)]{mathlibstats}
{Lean Prover Community}.
\newblock mathlib statistics.
\newblock \url{https://leanprover-community.github.io/mathlib_stats.html}, 2026.
\newblock Accessed: 2026-09-01.

\bibitem[Li et~al.(2025)Li, Liu, Chen, Lian, and Shao]{li2025reinforcedinformationretrieval}
Chaofan Li, Zheng Liu, Jianlyv Chen, Defu Lian, and Yingxia Shao.
\newblock Reinforced information retrieval, 2025.
\newblock URL \url{https://arxiv.org/abs/2502.11562}.

\bibitem[Lin et~al.(2025)Lin, Tang, Lyu, Yang, Chung, Zhao, Jiang, Geng, Ge, Sun, Wu, Gesi, Lu, Acuna, Yang, Lin, Choi, Chen, Arora, and Jin]{lin2025goedelproverv2scalingformaltheorem}
Yong Lin, Shange Tang, Bohan Lyu, Ziran Yang, Jui-Hui Chung, Haoyu Zhao, Lai Jiang, Yihan Geng, Jiawei Ge, Jingruo Sun, Jiayun Wu, Jiri Gesi, Ximing Lu, David Acuna, Kaiyu Yang, Hongzhou Lin, Yejin Choi, Danqi Chen, Sanjeev Arora, and Chi Jin.
\newblock Goedel-prover-v2: Scaling formal theorem proving with scaffolded data synthesis and self-correction, 2025.
\newblock URL \url{https://arxiv.org/abs/2508.03613}.

\bibitem[Liu et~al.(2026)Liu, Zhou, Zhu, Santos, He, Liu, Wang, Xie, Zhao, Wang, Zhi, Li, and Li]{liu2026numinaleanagentopengeneralagentic}
Junqi Liu, Zihao Zhou, Zekai Zhu, Marco~Dos Santos, Weikun He, Jiawei Liu, Ran Wang, Yunzhou Xie, Junqiao Zhao, Qiufeng Wang, Lihong Zhi, Jia Li, and Wenda Li.
\newblock Numina-lean-agent: An open and general agentic reasoning system for formal mathematics, 2026.
\newblock URL \url{https://arxiv.org/abs/2601.14027}.

\bibitem[Lu et~al.(2026)Lu, Emond, Yang, Chaudhuri, Sun, and Chen]{lu2026lean}
Jialin Lu, Kye Emond, Kaiyu Yang, Swarat Chaudhuri, Weiran Sun, and Wuyang Chen.
\newblock Lean finder: Semantic search for mathlib that understands user intents.
\newblock In \emph{International Conference on Learning Representations}, volume 2026, pp.\  155005--155040, 2026.

\bibitem[Moura \& Ullrich(2021)Moura and Ullrich]{moura2021lean}
Leonardo~de Moura and Sebastian Ullrich.
\newblock The {Lean} 4 theorem prover and programming language.
\newblock In \emph{CADE}, 2021.

\bibitem[{OpenAI}(2025)]{openai2025gptoss120bgptoss20bmodel}
{OpenAI}.
\newblock gpt-oss-120b and gpt-oss-20b model card, 2025.
\newblock URL \url{https://arxiv.org/abs/2508.10925}.

\bibitem[Ren et~al.(2025)Ren, Shao, Song, Xin, Wang, Zhao, Zhang, Fu, Zhu, Yang, et~al.]{ren2025deepseek}
ZZ~Ren, Zhihong Shao, Junxiao Song, Huajian Xin, Haocheng Wang, Wanjia Zhao, Liyue Zhang, Zhe Fu, Qihao Zhu, Dejian Yang, et~al.
\newblock Deepseek-prover-v2: Advancing formal mathematical reasoning via reinforcement learning for subgoal decomposition.
\newblock \emph{arXiv preprint arXiv:2504.21801}, 2025.

\bibitem[Shao et~al.(2024)Shao, Wang, Zhu, Xu, Song, Bi, Zhang, Zhang, Li, Wu, and Guo]{shao2024deepseekmathpushinglimitsmathematical}
Zhihong Shao, Peiyi Wang, Qihao Zhu, Runxin Xu, Junxiao Song, Xiao Bi, Haowei Zhang, Mingchuan Zhang, Y.~K. Li, Y.~Wu, and Daya Guo.
\newblock Deepseekmath: Pushing the limits of mathematical reasoning in open language models, 2024.
\newblock URL \url{https://arxiv.org/abs/2402.03300}.

\bibitem[Varambally et~al.(2026)Varambally, Voice, Sun, Chen, Yu, and Ye]{varambally2026hilbertrecursivelybuildingformal}
Sumanth Varambally, Thomas Voice, Yanchao Sun, Zhifeng Chen, Rose Yu, and Ke~Ye.
\newblock Hilbert: Recursively building formal proofs with informal reasoning, 2026.
\newblock URL \url{https://arxiv.org/abs/2509.22819}.

\bibitem[Wang et~al.(2017)Wang, Yu, Zhang, Gong, Xu, Wang, Zhang, and Zhang]{Wang_2017}
Jun Wang, Lantao Yu, Weinan Zhang, Yu~Gong, Yinghui Xu, Benyou Wang, Peng Zhang, and Dell Zhang.
\newblock Irgan: A minimax game for unifying generative and discriminative information retrieval models.
\newblock In \emph{Proceedings of the 40th International ACM SIGIR Conference on Research and Development in Information Retrieval}, SIGIR ’17, pp.\  515–524. ACM, August 2017.
\newblock \doi{10.1145/3077136.3080786}.
\newblock URL \url{http://dx.doi.org/10.1145/3077136.3080786}.

\bibitem[Wang et~al.(2026)Wang, Yao, Pan, Diao, and Zhang]{wang2026gargenerativeadversarialreinforcement}
Ruida Wang, Jiarui Yao, Rui Pan, Shizhe Diao, and Tong Zhang.
\newblock Gar: Generative adversarial reinforcement learning for formal theorem proving, 2026.
\newblock URL \url{https://arxiv.org/abs/2510.11769}.

\bibitem[Yang et~al.(2025)Yang, Li, Yang, Zhang, Hui, Zheng, Yu, Gao, Huang, Lv, Zheng, Liu, Zhou, Huang, Hu, Ge, Wei, Lin, Tang, Yang, Tu, Zhang, Yang, Yang, Zhou, Zhou, Lin, Dang, Bao, Yang, Yu, Deng, Li, Xue, Li, Zhang, Wang, Zhu, Men, Gao, Liu, Luo, Li, Tang, Yin, Ren, Wang, Zhang, Ren, Fan, Su, Zhang, Zhang, Wan, Liu, Wang, Cui, Zhang, Zhou, and Qiu]{yang2025qwen3technicalreport}
An~Yang, Anfeng Li, Baosong Yang, Beichen Zhang, Binyuan Hui, Bo~Zheng, Bowen Yu, Chang Gao, Chengen Huang, Chenxu Lv, Chujie Zheng, Dayiheng Liu, Fan Zhou, Fei Huang, Feng Hu, Hao Ge, Haoran Wei, Huan Lin, Jialong Tang, Jian Yang, Jianhong Tu, Jianwei Zhang, Jianxin Yang, Jiaxi Yang, Jing Zhou, Jingren Zhou, Junyang Lin, Kai Dang, Keqin Bao, Kexin Yang, Le~Yu, Lianghao Deng, Mei Li, Mingfeng Xue, Mingze Li, Pei Zhang, Peng Wang, Qin Zhu, Rui Men, Ruize Gao, Shixuan Liu, Shuang Luo, Tianhao Li, Tianyi Tang, Wenbiao Yin, Xingzhang Ren, Xinyu Wang, Xinyu Zhang, Xuancheng Ren, Yang Fan, Yang Su, Yichang Zhang, Yinger Zhang, Yu~Wan, Yuqiong Liu, Zekun Wang, Zeyu Cui, Zhenru Zhang, Zhipeng Zhou, and Zihan Qiu.
\newblock Qwen3 technical report, 2025.
\newblock URL \url{https://arxiv.org/abs/2505.09388}.

\bibitem[Yang et~al.(2023)Yang, Swope, Gu, Chalamala, Song, Yu, Godil, Prenger, and Anandkumar]{yang2023leandojo}
Kaiyu Yang, Aidan Swope, Alex Gu, Rahul Chalamala, Peiyang Song, Shixing Yu, Saad Godil, Ryan~J Prenger, and Animashree Anandkumar.
\newblock Leandojo: Theorem proving with retrieval-augmented language models.
\newblock \emph{Advances in Neural Information Processing Systems}, 36:\penalty0 21573--21612, 2023.

\bibitem[Zhang et~al.(2022)Zhang, Gong, Shen, Lv, Duan, and Chen]{AR22022ICLR}
Hang Zhang, Yeyun Gong, Yelong Shen, Jiancheng Lv, Nan Duan, and Weizhu Chen.
\newblock Adversarial retriever-ranker for dense text retrieval.
\newblock In \emph{The Tenth International Conference on Learning Representations, {ICLR} 2022, Virtual Event, April 25-29, 2022}, 2022.

\bibitem[Zhang et~al.(2026)Zhang, Xie, Sun, Chen, Wang, Li, Ding, and Zhou]{zhang2026onpolicyrlmeetsoffpolicy}
Wenhao Zhang, Yuexiang Xie, Yuchang Sun, Yanxi Chen, Guoyin Wang, Yaliang Li, Bolin Ding, and Jingren Zhou.
\newblock On-policy rl meets off-policy experts: Harmonizing supervised fine-tuning and reinforcement learning via dynamic weighting, 2026.
\newblock URL \url{https://arxiv.org/abs/2508.11408}.

\bibitem[Zhang et~al.(2025)Zhang, Li, Long, Zhang, Lin, Yang, Xie, Yang, Liu, Lin, Huang, and Zhou]{zhang2025qwen3embeddingadvancingtext}
Yanzhao Zhang, Mingxin Li, Dingkun Long, Xin Zhang, Huan Lin, Baosong Yang, Pengjun Xie, An~Yang, Dayiheng Liu, Junyang Lin, Fei Huang, and Jingren Zhou.
\newblock Qwen3 embedding: Advancing text embedding and reranking through foundation models, 2025.
\newblock URL \url{https://arxiv.org/abs/2506.05176}.

\end{thebibliography}
\bibliographystyle{iclr2027_conference}

\newpage
\appendix

\section{Extended Related Work}
\label{app:related_work}

\paragraph{Lean and Mathlib Retrieval.} Semantic search and retrieval for Lean and Mathlib have advanced rapidly, with a growing number of systems designed to help humans and agents locate relevant declarations in the library. LeanDojo~\citep{yang2023leandojo} introduced ReProver, which combines premise retrieval with tactic generation, while LeanExplore~\citep{asher2025leanexploresearchenginelean}, LeanDex~\citep{liu2026numinaleanagentopengeneralagentic}, and LeanSearch~\citep{gao2024semantic} support semantic search over declarations and input query rewriting. Lean Finder~\citep{lu2026lean} synthesized queries from user intents mined from Lean discussions and aligned retrieval with preference feedback. LeanSearch V2~\citep{gao2026leansearchv2globalpremise} used hierarchy-informed informalizations with embedding-based retrieval and reranking. These systems either train or simply embed using context-rich informalizations or synthetic data derived from Mathlib declarations, yet queries from humans or agents are usually short and ambiguous. None of the existing systems adapt to the real world ambiguous queries. Lean Finder V2 overcomes this limitation by scaling up generation of short and ambiguous queries with adversarial training, and optimizing the query-rewriting prompt to make implicit mathematical intent explicit before retrieval.

\paragraph{Adversarial and RL-Based Retrieval.} Reinforcement-learning-based approaches have increasingly been used to improve retrieval systems. Early work such as IRGAN~\citep{Wang_2017} formulated retrieval as a minimax game in which a generative model selects difficult documents for fixed queries to challenge a discriminative ranker, while AR2~\citep{AR22022ICLR} similarly co-trained a dense retriever and cross-encoder ranker through progressively harder negative documents. More recent work moves optimization to the query side. 
LeReT~\citep{hsu2024groundingtryingllmsreinforcement} and DeepRetrieval~\citep{jiang2025deepretrieval}  trained LLMs through RL to generate or rewrite queries that directly optimize retrieval metrics, while keeping the retriever fixed. Reinforced-IR~\citep{li2025reinforcedinformationretrieval} focused on generating hypothetical-document augmentations. \citet{jagatap2025reinforcement} used RL to generate synthetic product queries for classification models. Lean Finder V2 instead uses RL to generate concise and ambiguous queries that the current retriever fails on yet that remain faithful to their target declarations, and retrains the retriever on them.

\paragraph{Lean Provers and Agentic Systems.}
Recent advances in Lean theorem proving have improved formal reasoning through stronger provers and agentic system designs. DeepSeek-Prover-V2~\citep{ren2025deepseek} used recursive subgoal decomposition to construct cold-start reasoning data before reinforcement learning, while Goedel-Prover-V2~\citep{lin2025goedelproverv2scalingformaltheorem} combined scaffolded data synthesis with verifier-guided self-correction. GAR~\citep{wang2026gargenerativeadversarialreinforcement} further introduced an adversarial reinforcement-learning framework in which a problem composer adapts task difficulty to the evolving capability of a solver. Beyond standalone provers, systems such as Numina-Lean-Agent~\citep{liu2026numinaleanagentopengeneralagentic} and Hilbert~\citep{varambally2026hilbertrecursivelybuildingformal} integrated theorem proving with retrieval, decomposition, and verification, while research-level agents including Archon~\citep{ju2026automatedconjectureresolutionformal} and MechMathAgent~\citep{cao2026mechmathagentteamllm} orchestrated multiple reasoning components with the Lean compiler in the loop. These agents typically use infrastructure such as Lean-LSP-MCP~\citep{lean-lsp-mcp} to interact with the Lean compiler and auxiliary tools, including theorem-search engines for locating reusable Mathlib results. As a result, retrieval is becoming critical infrastructure for agentic formal reasoning. Lean Finder V2 focuses on improving this search infrastructure itself: it trains the retrieval model to better handle concise, ambiguous queries which are common from coding agents through adversarially generated supervision whose difficulty adapts to the retriever's evolving capability.

\begin{table}[h]
\centering
\captionsetup{font=small}
\caption{Median query length in tokens (Qwen3 tokenizer). Human and coding-agent queries are collected from 1 month period of Lean Finder usage.}
\label{tab:query-length}
{\small
\begin{tabular*}{\linewidth}{@{\extracolsep{\fill}}p{0.18\linewidth}p{0.48\linewidth}rr@{}}
\toprule
 & Source & Median length & Count \\
\midrule
\multirow{4}{=}{Training data} & Informalizations \citep{lu2026lean, gao2026leansearchv2globalpremise} & 82 & 309579 \\
 & Synthetic user queries \citep{lu2026lean} & 34 & 503378 \\
 & Our synthetic queries (Round 1) & 15 & 307085 \\
 & Our synthetic queries (Round 2) & 14 & 306948 \\
\midrule
\multirow{3}{=}{Real-world \& Test queries} & MathlibQR \citep{gao2026leansearchv2globalpremise} & 14 & 810 \\
 & Human queries & 16 & 20000 \\
 & Coding-agent queries & 15 & 100000 \\
\bottomrule
\end{tabular*}
}
\end{table}

\section{Ablation Studies}
\label{sec:ablations}

We conduct detailed ablation studies to answer two questions about adversarial query generation. First, do the alignment and difficulty rewards each contribute to effective synthetic data? Second, does another round of generator RL improve over generation from the base model, a fixed RL-trained generator, or simply sampling more queries? All comparisons use the same source declarations, context-rich data, retriever initialization, and matched retriever-update budgets. We evaluate both concise-query on MathlibQR~\citep{gao2026leansearchv2globalpremise} and context-rich benchmark~\citep{lu2026lean}.

\subsection{Contribution of Alignment and Difficulty Rewards}

\paragraph{Setup.}
We isolate the two RL reward components after one generator-retriever round.
The alignment-only, difficulty-only, and full-reward variants start from the same generator and Round 0 retriever, use matched generator-training
steps, and receive the same query-sampling budget. We then train their retrievers under matched training budgets. Round 0 adds no generated concise
queries and serves as the context-rich-only control. This setup isolates the effect of each reward component from generator compute, sampling volume, and retriever compute.

\begin{table}[h!]
\centering
\captionsetup{font=small}
\caption{Ablation of the query-generator reward after one generator-retriever
training round on the 810-query MathlibQR fair set. Round 0 uses only
context-rich data; the Round 1 variants differ only in the reward
used to train the query generator. Generator steps, sampling budgets, and
retriever steps are matched. Best result in each column is bold.}
\label{tab:reward-ablation}
{\small
\renewcommand{\arraystretch}{0.95}
\begin{tabular*}{0.9\linewidth}{@{\extracolsep{\fill}}lccc@{\hspace{1.25em}}ccc@{}}
\toprule
 & \multicolumn{3}{c}{nDCG@$k$} & \multicolumn{3}{c}{Recall@$k$} \\
\cmidrule(lr){2-4} \cmidrule(lr){5-7}
Configuration & 1 & 5 & 10 & 10 & 50 & 100 \\
\midrule
Round 0 & 0.463 & 0.590 & 0.611 & 0.760 & 0.864 & 0.901 \\
\midrule
Round 1 (difficulty only) & 0.453 & 0.575 & 0.602 & 0.762 & 0.870 & 0.919 \\
Round 1 (alignment only)  & 0.463 & 0.598 & 0.623 & 0.780 & 0.884 & 0.910 \\
\textbf{Round 1 (full reward)} & \textbf{0.520} & \textbf{0.657} & \textbf{0.682} & \textbf{0.842} & \textbf{0.925} & \textbf{0.951} \\
\bottomrule
\end{tabular*}
}
\end{table}

\begin{table}[h!]
\centering
\captionsetup{font=small}
\caption{Ablation of the query-generator reward after one generator-retriever
training round on the context-rich test set. The configurations
and matched training budgets are identical to
Table~\ref{tab:reward-ablation}. Best result in each column is bold.}
\label{tab:reward-ablation-information-rich}
{\small
\renewcommand{\arraystretch}{0.95}
\begin{tabular*}{0.9\linewidth}{@{\extracolsep{\fill}}lccc@{\hspace{1.25em}}ccc@{}}
\toprule
 & \multicolumn{3}{c}{nDCG@$k$} & \multicolumn{3}{c}{Recall@$k$} \\
\cmidrule(lr){2-4} \cmidrule(lr){5-7}
Configuration & 1 & 5 & 10 & 10 & 50 & 100 \\
\midrule
Round 0 & 0.563 & 0.720 & \textbf{0.743} & 0.910 & 0.967 & 0.981 \\
\midrule
Round 1 (difficulty only) & 0.530 & 0.689 & 0.711 & 0.879 & 0.956 & 0.975 \\
Round 1 (alignment only)  & 0.552 & 0.718 & 0.738 & 0.903 & 0.968 & 0.980 \\
\textbf{Round 1 (full reward)} & \textbf{0.569} & \textbf{0.728} & \textbf{0.743} & \textbf{0.916} & \textbf{0.985} & \textbf{0.993} \\
\bottomrule
\end{tabular*}
}
\end{table}

\paragraph{Results.}
The full reward is the only configuration that improves every MathlibQR metric over Round 0, and it outperforms both single-reward variants in every column (Table~\ref{tab:reward-ablation}). Difficulty alone degrades retrieval at the most important top-rank metrics and falls below Round 0. This pattern indicates that difficulty without alignment can reward queries that are hard for the wrong reason, producing mismatched query-declaration supervision that disrupts ranking. Alignment alone avoids this failure mode but yields only modest, uneven gains.

The context-rich results reinforce this conclusion
(Table~\ref{tab:reward-ablation-information-rich}). Alignment alone remains close to Round 0 across all metrics, whereas difficulty alone reduces
top rank metrics substantially (nDCG@10 from 0.743 to 0.711 and Recall@10 from 0.910 to 0.879). The full reward
instead achieves the strongest overall results across both query types: alignment keeps queries faithful to their declarations, and difficulty turns
those faithful queries toward current retriever gaps.

\subsection{RL-Trained Generation Versus Additional Sampling}

\paragraph{Setup.}
We next separate the effects of RL training, sampling volume, and an additional round of generator optimization. Every final retriever starts from the same base checkpoint, uses the same context-rich data, and receives a matched number of optimization steps. Round 0 adds no generated concise queries. In each branch, Round 1 adds one query per declaration in total and Round 2 uses two queries per declaration in total.

The frozen base-generator control samples both queries from Qwen3-4B before RL. The fixed RL-generator control samples both from the generator trained in the first RL round. Expert-queries SFT-generator control samples both from the generator trained with only SFT loss on the expert queries described in Section~\ref{subsec:query-generator}. Lean Finder V2 shares the same Round 1 generator, but applies another round of RL before generating the second query. We repeat the shared Round 1 result in the fixed-control and Lean Finder V2 table blocks so that their Round 2 outcomes can be compared directly.

\begin{table}[h!]
\centering
\captionsetup{font=small}
\caption{\textbf{Ablation of generator RL training
on the 810-query MathlibQR fair set.} Round 1 uses one generated query per
declaration, and Round 2 uses two in total. All retrievers receive a matched
number of optimization steps. Best result in each column is \textbf{bold}.}
\label{tab:retriever-adaptation-mathlibqr}
{\small
\renewcommand{\arraystretch}{0.95}
\begin{tabular*}{0.9\linewidth}{@{\extracolsep{\fill}}lccc@{\hspace{1.25em}}ccc@{}}
\toprule
 & \multicolumn{3}{c}{nDCG@$k$} & \multicolumn{3}{c}{Recall@$k$} \\
\cmidrule(lr){2-4} \cmidrule(lr){5-7}
Round & 1 & 5 & 10 & 10 & 50 & 100 \\
\midrule
Round 0 & 0.463 & 0.590 & 0.611 & 0.760 & 0.864 & 0.901 \\
\midrule
\multicolumn{7}{l}{\textit{Frozen base-generator control}} \\
Round 1 & 0.472 & 0.603 & 0.628 & 0.791 & 0.895 & 0.921 \\
Round 2 & 0.469 & 0.597 & 0.620 & 0.789 & 0.898 & 0.919 \\
\midrule
\multicolumn{7}{l}{\textit{Expert-query SFT-generator control}} \\
Round 1 & 0.475 & 0.610  & 0.632 & 0.794 & 0.904 & 0.919 \\
Round 2 & 0.473 & 0.607 & 0.633 & 0.795 & 0.896 & 0.914 \\
\midrule
\multicolumn{7}{l}{\textit{Fixed RL-generator control}} \\
Round 1 & 0.520 & 0.657 & 0.682 & 0.842 & 0.925 & 0.951 \\
Round 2 & 0.515 & 0.657 & 0.676 & 0.830 & 0.921 & 0.954 \\
\midrule
\multicolumn{7}{l}{\textit{Lean Finder V2}} \\
Round 1 & 0.520 & 0.657 & 0.682 & 0.842 & 0.925 & 0.951 \\
Round 2 & \textbf{0.558} & \textbf{0.707} & \textbf{0.732} & \textbf{0.893} & \textbf{0.956} & \textbf{0.973} \\
\bottomrule
\end{tabular*}
}
\end{table}

\begin{table}[h!]
\centering
\captionsetup{font=small}
\caption{\textbf{Ablation of generator RL training
on the context-rich test set.} The configurations are identical
to Table~\ref{tab:retriever-adaptation-mathlibqr}, and all retrievers receive a
matched number of optimization steps. Best result in each column is \textbf{bold}.}
\label{tab:retriever-adaptation-information-rich}
{\small
\renewcommand{\arraystretch}{0.95}
\begin{tabular*}{0.9\linewidth}{@{\extracolsep{\fill}}lccc@{\hspace{1.25em}}ccc@{}}
\toprule
 & \multicolumn{3}{c}{nDCG@$k$} & \multicolumn{3}{c}{Recall@$k$} \\
\cmidrule(lr){2-4} \cmidrule(lr){5-7}
Round & 1 & 5 & 10 & 10 & 50 & 100 \\
\midrule
Round 0 & 0.563 & 0.720 & 0.743 & 0.910 & 0.967 & 0.981 \\
\midrule
\multicolumn{7}{l}{\textit{Frozen base-generator control}} \\
Round 1 & 0.559 & 0.710 & 0.729 &0.894  & 0.967 & 0.981 \\
Round 2 & 0.552 & 0.705 & 0.725 & 0.890 & 0.965 & 0.981 \\
\midrule
\multicolumn{7}{l}{\textit{Expert-query SFT-generator control}} \\
Round 1 & 0.565 & 0.712 & 0.730 & 0.898 & 0.969 & 0.980 \\
Round 2 & 0.562  & 0.715 & 0.735 & 0.904 & 0.970 & 0.982 \\
\midrule
\multicolumn{7}{l}{\textit{Fixed RL-generator control}} \\
Round 1 & 0.569 & 0.728 & 0.743 & 0.916 & 0.985 & 0.993 \\
Round 2 & \textbf{0.573} & \textbf{0.729}  & 0.742 & 0.917 & 0.982 & 0.990 \\
\midrule
\multicolumn{7}{l}{\textit{Lean Finder V2}} \\
Round 1 & 0.569 & 0.728 & 0.743 & 0.916 & \textbf{0.985} & \textbf{0.993} \\
Round 2 & \textbf{0.573} & 0.728 & \textbf{0.749} &\textbf{0.919} & \textbf{0.985} & 0.992 \\
\bottomrule
\end{tabular*}
}
\end{table}

\paragraph{Results.}
RL training is more effective than generic query generation: from Round 0 to Round 1, it raises MathlibQR nDCG@10 from 0.628 to 0.682 and Recall@10 from 0.791 to 0.842 (Table~\ref{tab:retriever-adaptation-mathlibqr}). The SFT-only generator performs similarly to the frozen base generator across both rounds, with small top-rank gains, while the RL-trained generators achieve higher retrieval scores. Sampling a second query from any fixed generator adds no material gain, while applying another round of RL before generating the second query raises nDCG@10 to 0.732 and Recall@10 to 0.893. Lean Finder V2 Round 2 also achieves the best or close-to-best result in every context-rich metric (Table~\ref{tab:retriever-adaptation-information-rich}), preserving strong retrieval beyond concise queries.

\section{Detailed Evaluation Setup}
\label{app:detailed_evaluation_setup}

\subsection{Retrieval Evaluation}
\label{app:retrieval_evaluation_setup}

We submit the original benchmark query to every system and evaluate its returned retrievals against the ground-truth Mathlib declaration. For LeanExplore,
we obtain results from its public service at
\url{https://www.leanexplore.com/}. We omit LeanDex
\citep{liu2026numinaleanagentopengeneralagentic} from this comparison because we find that
it uses the LeanExplore retriever and adds query rewriting, which we evaluate
separately in Section~\ref{sec:gepa-results}. For LeanSearch V2, we run inference using
the authors' released code at
\url{https://github.com/frenzymath/LeanSearch-v2}, evaluating both its
retriever-only and retriever-plus-reranker configurations. For Lean Finder, we run inference using
the authors' released code at \url{https://github.com/delta-lab-ai/lean-finder}. For TheoremGraph~\citep{kurgan2026theoremgraphbridgingformalinformal}, we report the numbers found in \citet{kurgan2026theoremgraphbridgingformalinformal}, and since its evaluation uses a cutoff of 10, Recall@50 and Recall@100 are not reported. 

For the general-agent baseline, we use Claude Code with Sonnet 5 and permit
online search over Mathlib documentation and source pages. We explicitly
disallow Lean search-engine websites and tools so that the agent cannot delegate
retrieval to one of the specialized systems under comparison. To control cost,
we process ten independent queries in each Claude Code invocation, and require
a separate ranked list of exactly ten Mathlib declaration names for every
query. Ten candidates provide a fixed top-10 retrieval that supports
direct computation of nDCG@1, nDCG@5, nDCG@10, and Recall@10. Because the agent
returns no candidates beyond rank 10, we do not report Recall@50 or Recall@100
for Claude Code. The cost of running Claude Code with Sonnet 5 is approximately \$60 for the two benchmarks.

\subsection{Query-Rewriting Evaluation}
\label{app:query_rewriting_evaluation_setup}

All rewriting methods receive the same original MathlibQR query, and their rewritten
output replaces that original query as the input to Lean Finder V2. GEPA is initialized
with the seed prompt provided in Appendix~\ref{app:seed_prompt}.
This seed is the short, pre-optimization instruction that initializes GEPA and
defines the rewriting task. The GEPA-optimized prompt is provided in Appendix~\ref{app:gepa-prompt}. We obtain the
LeanSearch augmentation from its public
augmentation endpoint at \url{https://leansearch.net/augment}. For LeanDex, we
send the query to \url{https://leandex.projectnumina.ai/api/v1/search} using
the \texttt{q} parameter and \texttt{generate\_query=true}, then use the
returned \texttt{agent\_query} as the rewrite. Each resulting query is evaluated
against the same ground-truth declaration and with the same fixed Lean Finder
V2 model. For every rewriting method, we independently generate the complete
rewritten query set three times and report the mean and standard deviation of
the retrieval metrics across these runs. The no-rewrite condition bypasses
generation and is evaluated once on the original queries.

\subsection{Downstream Theorem-Proving Evaluation}
\label{app:theorem_proving_evaluation_setup}

FATE-H contains 100 challenging formal algebra problems drawn from research
literature, PhD qualifying examinations, and textbooks. We evaluate all five
retrieval conditions on these problems using Claude Code with Sonnet 5. Each
retrieval engine is exposed to the agent
through the same tool interface, while the agent prompt and execution
environment remain fixed. We impose a \$1 token-cost cap on every problem for
all conditions due to cost. The agent is not permitted to change the original theorem
statement. We count a problem as passed only when the produced Lean file is
sorry-free, compiles in the benchmark environment, and proves the original
statement within the cost cap.

We perform three independent runs for each retrieval condition and report the
mean pass rate and sample standard deviation in percentage points across runs. Across a single complete 100-problem run, token costs are approximately
\$90 with Lean Finder V2, \$92 with LeanSearch V2, \$94 with no semantic search tool, \$97 with Lean Finder, and \$98 with LeanExplore.

\section{LLM-Judge Evaluation and Validation}
\label{app:llm-judge-validation}

\paragraph{Deployment-query protocol.}
We sample 200 queries from interactions with Lean search engines, covering
queries issued by both humans and coding agents. Unlike MathlibQR, these queries
do not come with a designated target declaration. For every query, Lean Finder
V2, LeanSearch V2, Lean Finder, LeanExplore, and Claude Code each return five
declarations. To ensure fair comparison, we randomly order the five systems' retrievals for every query, and mask the identities of each system. GPT-5.6 Sol and Claude Opus 5 independently inspect the query and
the five result lists, then rank the systems from most to least helpful. We average the ordinal ranks across queries and separately
report how often each system is ranked first. The two judges choose the same
rank-1 system on 89.0\% of queries, showing that the preference for
Lean Finder V2 is consistent across judge families.

\paragraph{Validation of LLM judge reliability against human-curated labels.}
We validate the same judges on all 810 queries in the MathlibQR fair set
\citep{gao2026leansearchv2globalpremise} to ensure the reliability of LLM judges. For each query, we construct a five-way choice from the top Lean Finder V2 retrievals and ensure that the
human-labeled declaration is present. When it is absent, we replace one retrieved declaration with the ground truth; the five choices are then shuffled. Each judge sees only the query and each choice's formal name, declaration kind, and Lean type. It does not see the benchmark label, original retrieval rank, retrieval score, or which choice was inserted. Agreement is the fraction of queries for which the judge selects the human-labeled declaration. We report the result in Table~\ref{tab:judge-label-validation}.

\begin{table}[h]
\centering
\captionsetup{font=small}
\caption{Agreement of GPT-5.6 Sol and Claude Opus 5 with human-curated labels
on the 810-query MathlibQR fair set. Each judge selects one of five candidate
declarations; we report agreement counts and rates.}
\label{tab:judge-label-validation}
{\small
\renewcommand{\arraystretch}{0.95}
\begin{tabular}{@{}lcc@{}}
\toprule
Judge & Agreements & Agreement (\%) \\
\midrule
GPT-5.6 Sol & 738/810 & 91.11 \\
Claude Opus 5 & 745/810 & 91.98 \\
\bottomrule
\end{tabular}
}
\end{table}

Both judges recover the human-selected MathlibQR declaration on more than 91\% of queries despite choosing among the retriever's strongest confusions.
This independent agreement supports the reliability of the preference evaluation conducted with the same LLM judges. It also provides a second validation of MathlibQR itself: human
curation and two frontier model families converge on the same labels, making widespread false-positive benchmark labels unlikely.

\section{Reranker Training and Results}
\label{app:reraker_results}

In this section, we study reranking on top of our strong Lean Finder V2 retrieval model. 
We fine-tune Qwen3-Reranker-8B to estimate the relevance of Mathlib declarations to a given query. The queries used for training include the large-scale context-rich queries described in Section~\ref{subsec:base-retriever} as well as the synthetic short queries described in Section~\ref{subsec:query-generator}. For each query, we use the ground-truth declaration as a positive example and randomly sample negative examples from the remaining Mathlib declarations. The model jointly processes each query-declaration pair and predicts its relevance using the output logits, $z_{\texttt{yes}}$ and $z_{\texttt{no}}$, for the \texttt{yes} and \texttt{no} tokens. Specifically, the relevance score is
\begin{equation}
    s_\theta(q,d)
    = \sigma\!\left(z_{\texttt{yes}} - z_{\texttt{no}}\right),
\end{equation}
where $\sigma$ denotes the sigmoid function. We minimize binary cross-entropy over the training pairs:
\begin{equation}
    \mathcal{L}(\theta)
    = -\frac{1}{|\mathcal{D}|}
    \sum_{(q,d,y)\in\mathcal{D}}
    \left[
        y \log s_\theta(q,d)
        + (1-y)\log\!\left(1-s_\theta(q,d)\right)
    \right],
\end{equation}
where $y=1$ indicates a positive pair and $y=0$ indicates a sampled negative.

\begin{table}[h!]
\centering
\captionsetup{font=small}
\caption{Retrieval performance on the 810-query MathlibQR fair set of short,
human-written queries. Systems include LeanExplore, LeanSearch V2, Claude Code
(Sonnet 5), Lean Finder, Lean Finder V2 with and without our reranker, and Lean Finder V2 with both our reranker and GEPA-optimized query rewriting.
Lean Finder V2 denotes the retriever-only model. Claude Code returns only ten
candidates, so its Recall@50 and Recall@100 are not reported. Best results in
each column are bold.}
\label{tab:mathlibqr-retrieval-reranker}
{\small
\renewcommand{\arraystretch}{0.95}
\begin{tabular*}{\linewidth}{@{\extracolsep{\fill}}lccc@{\hspace{1.25em}}ccc@{}}
\toprule
 & \multicolumn{3}{c}{nDCG@$k$} & \multicolumn{3}{c}{Recall@$k$} \\
\cmidrule(lr){2-4} \cmidrule(lr){5-7}
System & 1 & 5 & 10 & 10 & 50 & 100 \\
\midrule
LeanExplore                         & 0.246 & 0.358 & 0.393 & 0.569 & 0.743 & 0.789 \\
LeanSearch V2 (retriever-only)      & 0.340 & 0.472 & 0.494 & 0.657 & 0.790 & 0.830 \\
LeanSearch V2 (retriever+reranker)    & 0.470 & 0.601 & 0.623 & 0.780 & 0.847 & 0.858 \\
Claude Code (Sonnet 5)              & 0.527 & 0.629 & 0.641 & 0.747 & -- & -- \\
Lean Finder                      & 0.370 & 0.514 & 0.533 & 0.698 & 0.824 & 0.875 \\
Lean Finder V2          &      0.558 & 0.707 & 0.732 &  \textbf{0.893} & 0.956   &   0.973    \\
Lean Finder V2 (w/ reranker)            &      0.575 & 0.714 & 0.735 &  0.890 & 0.968   &   0.980    \\
Lean Finder V2 (w/ reranker+gepa rewrite)            &      \textbf{0.610} & \textbf{0.726} & \textbf{0.746} &  0.890 & \textbf{0.970}   &   \textbf{0.981}    \\
\bottomrule
\end{tabular*}
}
\end{table}

\begin{table}[h!]
\centering
\captionsetup{font=small}
\caption{Retrieval performance on the 2{,}526-query context-rich
test set. Systems in comparison include LeanExplore, LeanSearch V2, Claude Code (Sonnet 5),
Lean Finder, and Lean Finder V2 with and without our reranker. Lean Finder V2
denotes the retriever-only model. Claude Code returns only ten candidates, so
its Recall@50 and Recall@100 are not reported. Best results in each column are
bold.}
\label{tab:information-rich-retrieval-reranker}
{\small
\renewcommand{\arraystretch}{0.95}
\begin{tabular*}{0.9\linewidth}{@{\extracolsep{\fill}}lccc@{\hspace{1.25em}}ccc@{}}
\toprule
 & \multicolumn{3}{c}{nDCG@$k$} & \multicolumn{3}{c}{Recall@$k$} \\
\cmidrule(lr){2-4} \cmidrule(lr){5-7}
System & 1 & 5 & 10 & 10 & 50 & 100 \\
\midrule
LeanExplore                         & 0.180 & 0.376 & 0.406 & 0.661 & 0.837 & 0.872\\
LeanSearch V2 (retriever-only)      & 0.414 & 0.569 & 0.596 & 0.783 & 0.907 & 0.942 \\
LeanSearch V2 (retriever+reranker)    & 0.506 & 0.669 & 0.691 & 0.873 & 0.954 & 0.964 \\
Claude Code (Sonnet 5)    & 0.245 & 0.345 & 0.363 & 0.485 & -- & -- \\
Lean Finder                      & 0.544 & 0.720 & 0.742 & 0.914 & 0.980 & 0.989 \\
Lean Finder V2                 & 0.573 & 0.728 & 0.749 & 0.919 & 0.985 & 0.992 \\
Lean Finder V2 (w/ reranker)                 & \textbf{0.607} & \textbf{0.760} & \textbf{0.779} & \textbf{0.939} & \textbf{0.987} & \textbf{0.995} \\
\bottomrule
\end{tabular*}
}
\end{table}

Table~\ref{tab:mathlibqr-retrieval-reranker} and Table~\ref{tab:information-rich-retrieval-reranker} show that adding the reranker further improves the ranking quality on top of Lean Finder V2's strong performance on both MathlibQR and context-rich queries. With the GEPA-optimized query rewriting and reranking both applied to Lean Finder V2, we achieve even more improvements across all metrics (except for Recall@10), reaching 0.610 for nDCG@1 on MathlibQR.

\section{Training Details}
\label{app:training_details}

We describe the training details of our generator, retriever, reranker, and our prompt optimization process in this section. All experiments are conducted on NVIDIA 6000Ada and L40S GPUs.

\subsection{Generator Training Details}
\label{app:generator_training_details}
We initialize the generator from Qwen3-4B
\citep{yang2025qwen3technicalreport} and use GPT-OSS-120B
\citep{openai2025gptoss120bgptoss20bmodel} as the alignment judge. We apply
GRPO to $2{,}000$ Mathlib statements in each round. We train for
one epoch with learning rate $5\times10^{-7}$, group size $G=8$,
sampling temperature 1.0, $\beta=0.001$, and a 1\% warmup. The first RL round additionally mixes expert
deployment queries through the CHORD objective. Here, $\mu$ has
peak value 0.9 and valley value 0.05, with no warmup and 30 decay steps. We
use the expert data only in this first round because the two supervision
sources serve different roles. The expert queries provide a prior over the
phrasing, brevity, and search patterns observed in deployment, which the rank
and alignment rewards are not designed to encode. After the first round, the
generator enters the next round from this deployment-aligned checkpoint, while
the updated retriever defines a new difficulty landscape. We therefore remove
the fixed expert objective in the second round and let the on-policy reward
adapt the generator to the updated retriever. The alignment gate and length
penalty remain active, preserving target fidelity and concise query form during this adaptation. Each round generates over 300K queries, and the final combined synthetic query-declaration pairs from adversarial training reaches over 600K.

We now define our query overlong penalty $p_{\mathrm{len}}(L)$, which is used as part of the RL reward function in Section~\ref{subsec:query-generator}. For query length $L$, measured with the retriever tokenizer, the soft length
penalty is
\begin{equation}
  p_{\mathrm{len}}(L)=
  \begin{cases}
    0, & L\leq L_{\max},\\
    \dfrac{L-L_{\max}}{L_{\mathrm{buf}}},
      & L_{\max}<L\leq L_{\max}+L_{\mathrm{buf}},\\
    1, & L>L_{\max}+L_{\mathrm{buf}}.
  \end{cases}
  \label{eq:length-penalty}
\end{equation}
We set $L_{\max}=40$, $L_{\mathrm{buf}}=10$, and
$\lambda_{\mathrm{len}}=0.15$. The penalty is zero through 40 tokens, grows
linearly from 40 to 50 tokens, and remains at its maximum thereafter.
Appendix~\ref{app:reward_hyperparameters} motivates these settings using the
query-length distributions of human users and coding agents.

For the expert dataset used in SFT component of the hybrid objective, we use records from human and coding-agents interactions with current Lean search engines through lean-lsp-mcp~\citep{lean-lsp-mcp}, including Lean Finder~\citep{lu2026lean} and  LeanSearch~\citep{gao2024semantic, gao2026leansearchv2globalpremise}. Since those queries don't have ground truth label, we use an LLM-assisted label discovery process. We first use an existing search engine (Lean Finder) to retrieve the top 50 Mathlib declarations for the real world queries, and employ an LLM-judge (Opus 5) to determine the correct ground truth Mathlib declaration. If no ground truth is found in the top 50 retrievals, then we skip the query. We keep a total of 2{,}000 query-declaration pairs as the expert dataset. The reliability of Opus 5 as the LLM judge in replicating human preference on quality of retrieval is validated in Appendix~\ref{app:llm-judge-validation}, and previous research showed that LLM judges closely match human judgments of which retrieved Mathlib statements are useful~\citep{lu2026lean}. Using a frontier LLM judge is the only practical way to reliably scale the data preparation process. We also filter the expert queries used for SFT against the 200 held-out queries used for LLM-judge-based evaluation (Section~\ref{sec:deployment-preference}) and all other datasets used in evaluation (context-rich queries and MathlibQR), using exact-match deduplication and MinHash-LSH-based near-duplicate detection to ensure fairness.

\subsection{Retrieval Training Details}
\label{app:retrieval_training_details}
We initialize our retriever with Qwen3-Embedding-8B
\citep{zhang2025qwen3embeddingadvancingtext}, and fine-tune using masked
InfoNCE. We use a per-device batch size of 32, learning rate $10^{-5}$, 5\%
warmup, $\tau=0.1$, and $m=0.1$. Each
round combines the context-rich data with the accumulated concise queries generated up to that round. For a controlled comparison, all retriever variants receive matched optimization steps.

We show the masked InfoNCE loss here.
Let $B$ denote the mini-batch size, $q_i$ the $i$-th query, and $c_i^+$ its
ground-truth positive declaration. The candidate set
$\mathcal{C}_B = \{c_j^+\}_{j=1}^{B}$ contains all positive declarations in
the mini-batch. We write $\mathrm{Sim}(\cdot,\cdot)$ for cosine similarity,
$\tau$ for the softmax temperature, and $m$ for the false-negative margin. The
indicator $\mathbb{I}(q_i,c)$ is zero when
$\mathrm{Sim}(q_i,c) > \mathrm{Sim}(q_i,c_i^+) + m$, masking a candidate that
scores higher than the labeled positive by more than $m$, and one otherwise.
The masked InfoNCE objective is
\begin{equation}
\mathcal{L}_{\mathrm{ret}}
= -\frac{1}{B}\sum_{i=1}^{B}
\log
\frac{\exp\!\left(\mathrm{Sim}(q_i,c_i^+)/\tau\right)}
{\sum_{c \in \mathcal{C}_B}
\mathbb{I}(q_i,c)\exp\!\left(\mathrm{Sim}(q_i,c)/\tau\right)}.
\label{eq:masked_infonce}
\end{equation}

\subsection{Reranker Training Details}
\label{app:reranker_training_details}
We train the reranker models for one epoch using LoRA \citep{lora} (rank $r=8$, $\alpha=32$) on all linear layers, with a learning rate of $5\times10^{-5}$, a warmup ratio of $0.05$, and a per-device batch size of 64. 

\subsection{Prompt Optimization Training Details}
\label{app:prompt_opt_details}

\paragraph{Optimization procedure.}
We optimize the rewriting prompt using GEPA~\citep{agrawal2026gepareflectivepromptevolution}.
Starting from the seed rewriting prompt, the optimizer maintains a pool of candidate prompts.
The training data are split into a feedback set
$\mathcal{D}_{\mathrm{feedback}}$ and a Pareto set
$\mathcal{D}_{\mathrm{pareto}}$.
At each optimization step, GEPA samples a candidate prompt that is best-performing on at least one example in $\mathcal{D}_{\mathrm{pareto}}$, and evaluates it on a minibatch from $\mathcal{D}_{\mathrm{feedback}}$.
The resulting scalar scores and textual retrieval-grounded feedback are provided to a reflection model, which proposes a revised prompt. More details on the scalar score and the retrieval-grounded feedback are provided in the following two paragraphs.
The revision is retained only if it improves the average minibatch score; retained candidates are then evaluated on the $\mathcal{D}_{\mathrm{pareto}}$ and added to the candidate pool.
Candidate selection favors non-dominated prompts that achieve the best score on one or more examples in $\mathcal{D}_{\mathrm{pareto}}$, with higher sampling probability for candidates that are best on more examples.
After the optimization budget is exhausted, we return the candidate with the highest average score on
$\mathcal{D}_{\mathrm{pareto}}$.

\paragraph{Scalar score.}
Let
$(q_i,d_i)$
denote each query-declaration pair used for prompt optimization.
For a rewriting prompt $p$, let $rank_i(p)$ denote the rank of the target declaration $d_i$ when the fixed retriever receives the rewrite of $q_i$ under $p$;
$rank_i^{\mathrm{orig}}$ and $rank_i^{\mathrm{seed}}$ denote its ranks for the original query and seed-prompt rewrite, respectively.
For each example, we return the scalar score
\begin{equation}
\label{eq:rewrite_score}
s_i(p)
=
\lambda_{\mathrm{abs}}
\frac{1}{\log_2(rank_i(p)+1)}
+
\lambda_{\mathrm{orig}}
\frac{rank_i^{\mathrm{orig}}}
     {rank_i^{\mathrm{orig}}+rank_i(p)}
+
\lambda_{\mathrm{seed}}
\frac{rank_i^{\mathrm{seed}}}
     {rank_i^{\mathrm{seed}}+rank_i(p)} .
\end{equation}
The first term rewards high absolute target rank, while the latter two reward improvement over the original query and the seed-prompt rewrite.
We set
$\lambda_{\mathrm{abs}}=0.5$
and
$\lambda_{\mathrm{orig}}=\lambda_{\mathrm{seed}}=0.25$,
placing greater emphasis on absolute retrieval quality while weighting improvements over the two baselines equally.

\paragraph{Retrieval-grounded feedback.}
The natural-language diagnostic feedback for each rewrite contains the target Lean declaration and its informal description, the original query, the current rewrite, the original and current target ranks, the seed-prompt rewrite and its target rank, the scalar score, and the top-$10$ retrieved declarations.
We also provide principles of reflection, which ask the reflection model to compare the current, original, and seed outcomes, analyze both gains and regressions, and propose a reusable behavioral change. 

\paragraph{Hyperparameters.}
We use 200 training examples, split evenly between
$\mathcal{D}_{\mathrm{feedback}}$
and
$\mathcal{D}_{\mathrm{pareto}}$,
so that
$|\mathcal{D}_{\mathrm{feedback}}|
=
|\mathcal{D}_{\mathrm{pareto}}|
=
100$.
We use a reflection minibatch size of $3$ and an optimization budget of $B= 1{,}000$, representing the maximum number of scalar score evaluations.
Here, the minibatch size determines how many feedback examples are used to construct each reflective prompt update, while the Pareto-set size determines how many examples are used to evaluate and select prompt candidates.

\paragraph{Training data.}
We construct the optimization data from a subset of the concise query-declaration pairs produced by the RL-trained generator in Section~\ref{subsec:query-generator}.
Each optimization example contains the original query, the target declaration, its rank based on the original query, the seed-prompt-rewritten query as well as its retrieval rank.

\subsection{Pseudocode for Training Process}

\begin{algorithm}[H]
  \caption{Retriever-based adversarial training}
  \label{alg:training-loop}
  \begin{algorithmic}[1]
    \Require Mathlib declarations $\mathcal D$, context-rich query data
      $\mathcal D_{\mathrm{long}}$, SFT anchor data
      $\mathcal D_{\mathrm{SFT}}$, SFT mixing schedule $\{\mu_t\}$, base generator $\pi_{\theta_0}$,
      number of rounds $T$
    \State $R_0 \gets
      \Call{TrainRetriever}{\mathcal D_{\mathrm{long}}}$
    \For{$t=0,\ldots,T-1$}
      \State $\theta_{t+1}\gets
        \Call{HybridObjective}{\theta_t,R_t,\mathcal D,\mathcal D_{\mathrm{SFT}},\mu_t}$
      \State $\mathcal D_{t+1}^{\mathrm{short}}\gets\emptyset$
      \ForAll{$d\in\mathcal D$}
        \State $q\sim\pi_{\theta_{t+1}}(\cdot\mid d)$
        \State $\mathcal D_{t+1}^{\mathrm{short}}\gets
          \mathcal D_{t+1}^{\mathrm{short}}\cup\{(q,d)\}$
      \EndFor
      \State $\mathcal D_{\mathrm{short}}\gets
        \mathcal D_{\mathrm{short}}\cup\mathcal D_{t+1}^{\mathrm{short}}$
      \State $R_{t+1}\gets\Call{TrainRetriever}{
        \mathcal D_{\mathrm{long}}\cup\mathcal D_{\mathrm{short}}}$
    \EndFor
    \State \Return $R_T$
  \end{algorithmic}
\end{algorithm}

Here $\mu_t$ is the SFT mixing schedule of round $t$, as specified in Appendix~\ref{app:generator_training_details}; when $\mu_t=0$, \textsc{HybridObjective} reduces to the on-policy GRPO objective.

\section{Model Scale and Backbone Choice}
\label{app:model_scale}

Table~\ref{tab:model-scale} compares the parameter counts of the retrieval
systems we evaluated. Lean Finder V2 uses a
single 8B retriever, while the reranked LeanSearch V2 configuration combines an
8B retriever with an 8B reranker.

\begin{table}[h!]
\centering
\captionsetup{font=small}
\caption{\textbf{Retrieval-model scale for the systems in
Table~\ref{tab:mathlibqr-retrieval}.} For a two-stage system, the values before
and after $+$ give the retriever and reranker sizes, respectively.}
\label{tab:model-scale}
\begin{tabular}{lcc}
\toprule
System & Retrieval configuration & Parameters \\
\midrule
LeanExplore & Retriever & 109M \\
LeanSearch V2 & Retriever only & 8B \\
LeanSearch V2 & Retriever + reranker & 8B + 8B \\
TheoremGraph & Retriever & 8B \\
Lean Finder & Retriever & 7B \\
Lean Finder V2 & Retriever & 8B \\
\bottomrule
\end{tabular}
\end{table}

Lean Finder V2 and LeanSearch V2 rely on the same Qwen3-Embedding-8B base checkpoint~\citep{gao2026leansearchv2globalpremise,zhang2025qwen3embeddingadvancingtext}.
LeanSearch V2 applies this checkpoint without task-specific retriever fine-tuning, whereas Lean Finder V2 fine-tunes it for Mathlib retrieval. The
shared initialization shows that our gains come from adapting the same backbone to Mathlib search rather than increasing retriever scale.

Our method is agnostic to the base model architecture and pretrained checkpoint. Although we can ablate other base models, our goal is to deliver a strong Mathlib search engine for
the AI4Math and Lean communities so starting from a strong base model is necessary.

\section{Reward Hyperparameter Justifications}
\label{app:reward_hyperparameters}

\subsection{Length-penalty thresholds}
The length penalty guards against unnecessarily long generations rather than
forcing queries toward a fixed target length. In the deployment data summarized
in Figure~\ref{fig:motivation}(a), the 90th-percentile lengths of distinct human
and coding-agent queries are 37 and 28 tokens, respectively, while their
95th-percentile lengths are 45 and 34 tokens. We set $L_{\max}=40$, above the
90th percentile for both sources, so that queries in the ordinary deployment
range remain unpenalized. We use $L_{\mathrm{buf}}=10$ so that the penalty grows
smoothly from 40 to 50 tokens and reaches its maximum only beyond the 95th
percentile of both sources. Finally, $\lambda_{\mathrm{len}}=0.15$ caps the
length deduction at 0.15 on the unit-scale difficulty reward, keeping length as
a secondary regularizer. These choices preserve natural variation in query
length while discouraging the generator from adding unnecessary text to lower
the target declaration's rank.

\subsection{Rank regions for the difficulty reward}

We calibrate the rank regions to the retrieval depths used in deployment rather
than to downstream benchmark performance. Our logs cover queries from human users and coding agents to deployed Lean
search engines, each with a valid requested retrieval depth. Among these
requests, 99.9\% of human requests and 98.7\% of coding-agent requests
return at least five results. A target within the top five is therefore almost
always exposed to the downstream user or agent. We treat such queries as
already solved by the retriever and assign no difficulty reward for
$k\leq5$, directing generation toward queries that reveal unresolved retrieval
gaps.

We set $k_{\mathrm{sat}}=30$ because 100\% of human requests and 99.9\% of
coding-agent requests return at most 30 results. The logarithmic ramp from ranks
6 to 30 rewards increasingly challenging queries, while saturation beyond rank
30 treats all queries outside the effective deployment range equally. This
plateau rewards the generator for exposing a meaningful retrieval gap without
incentivizing it to push the target arbitrarily farther down the ranking by
removing target information.

We set $k_{\mathrm{cut}}=100$ as a complementary retrieval-side safeguard. Queries that place the source declaration beyond rank 100 tend to
omit target-identifying content so we assign zero difficulty reward for
$k>100$, keeping optimization focused on challenging queries that remain aligned to their target declarations.

Since each boundary changes the on-policy generator
and consequently the retriever's training data, testing another setting requires
retraining the complete generator-retriever pipeline. We therefore do not select these thresholds through
a benchmark-driven hyperparameter sweep, as the sweep would cost significant GPU resources. The synthetic data produced by the
generator trained with this fixed setting are already useful and show improvements across complementary
model-side training (Table~\ref{tab:mathlibqr-retrieval} and Table~\ref{tab:information-rich-retrieval}) and input-side prompt optimization (Table~\ref{tab:gepa-rewriting}), without even performing careful tuning.

\section{Alignment-Gate Audit}
\label{app:alignment_audit}
We audit alignment-gate reliability on 100 query-declaration pairs picked
from generator outputs across the two training rounds to reach a balanced negative and positive pair distribution for effective auditing (40 pairs are misaligned and 60 pairs are aligned). GPT-OSS-120B agrees
with the human labels on 98/100 pairs and with Claude Opus 5 on 97/100 pairs.
It incorrectly accepts 1/40 human-labeled misaligned pairs, corresponding to a
2.5\% false-positive rate, and rejects 1/60 human-labeled aligned pairs, giving
a 1.7\% false-negative rate. These results demonstrate that false alignment
approvals are rare while the gate preserves valid training queries.

\section{Query Rewriting Case Study}
\label{app:rewrite_example}

We consider a query whose target declaration is
\begin{center}
\texttt{CategoryTheory.Adjunction.homEquiv\_naturality\_left}.
\end{center}
The original query states the desired equation but leaves the role of
\(\Phi\) implicit. Below are the complete query texts for one generation
of each rewriting method.

\begin{rewritecase}{Original (rank 7).}\small
For \(F \dashv G\),
\(\Phi(F.\mathrm{map}\,f \gg g)=f\gg\Phi(g)\).
\end{rewritecase}

\begin{rewritecase}{GEPA-optimized prompt rewriting (rank 1).}\small
For an adjunction \(F \dashv G\) and its hom-set equivalence
\(\Phi\) (Lean \texttt{Adjunction.homEquiv}), the equation
\(\Phi(F.\mathrm{map}\,f \gg g)=f\gg\Phi(g)\)
(left naturality of the adjunction hom-equivalence).
\end{rewritecase}

\begin{rewritecase}{LeanSearch query rewriting (rank 8).}\small
Adjunction Naturality (Left Triangle Identity): Given two functors
\(F\) and \(G\) that form an adjunction with \(F\) left adjoint to
\(G\), and given the natural bijection between morphisms from \(F\)
applied to an object to another object and morphisms from the
original object to \(G\) applied to that other object, the natural
bijection applied to the composition of \(F\) applied to a morphism
followed by another morphism equals the composition of that first
morphism followed by the natural bijection applied to the second
morphism.
\end{rewritecase}

\begin{rewritecase}{LeanDex query rewriting (rank 16).}\small
The property
\(\Phi(F.\mathrm{map}\,f \gg g)=f\gg\Phi(g)\)
for an adjunction \(F \dashv G\) relating natural transformations
and functor maps within category theory.
\end{rewritecase}

\begin{rewritecase}{Seed prompt query rewriting (rank 7)}\small
Lemma that for an adjunction \(F \dashv G\),
\(\Phi(F.\mathrm{map}\,f \gg g)=f\gg\Phi(g)\).
\end{rewritecase}

Mathlib's theorem states precisely that the adjunction hom-set
equivalence sends \(F.\mathrm{map}\,f \gg g\) to
\(f\gg\Phi(g)\). GEPA preserves this equation and its adjunction
setting while supplying one useful clarification: \(\Phi\) is
\texttt{Adjunction.homEquiv}, and the requested relationship is its
\emph{left naturality}. This follows the optimized prompt's principles
to preserve mathematical intent, keep the exact relationship in focus,
and use standard Lean terminology only when it directly denotes that
relationship. GEPA introduces no additional assumptions or adjacent
theorem.

The alternatives illustrate why more text is not necessarily better.
LeanSearch augmentation replaces the exact equation with a lengthy paraphrase and
incorrectly labels it a ``Left Triangle Identity''; Mathlib's triangle
identities instead concern the adjunction's unit and counit. LeanDex augmentation
adds broad natural-transformation context without identifying the
hom-set equivalence, shifting attention toward other adjunction
declarations. The seed rewrite adds only ``lemma that,'' leaving the
missing context unresolved. GEPA's focused rewrite raises the target
from rank 7 to rank 1, whereas these representative alternatives rank
it 8, 16, and 7, respectively.

\section{Prompts}
\label{app:prompts}

\subsection*{Seed Prompt for Query Rewriting}
\label{app:seed_prompt}
\begin{promptbox2}
\ttfamily\small
Rewrite the user's search query for Mathlib4.\\ \\
original query: ...

\medskip
{\normalfont\bfseries OUTPUT FORMAT:}

rewritten\_query: <the rewritten query>
\end{promptbox2}

\subsection*{Optimized Prompt for Query Rewriting}
\label{app:gepa-prompt}

\begin{promptbox2}
\ttfamily\small

You are an expert in mathematics, Lean 4, and theorem retrieval.

\medskip
{\normalfont\bfseries INSTRUCTION:}

Rewrite the user's search query for mathematical/Lean theorem retrieval.

\medskip
{\normalfont\bfseries REWRITING PRINCIPLES:}

1. Preserve the original mathematical intent, setting, assumptions, and ambiguity.

\smallskip
2. If the query already clearly names the object and relationship, keep it nearly verbatim.

\smallskip
3. Otherwise, expand terse terms only with standard synonyms or equivalent formulations directly supported by the query, preferring one concise clarification over stacked paraphrases or alternative formulations.

...\\ \\
original query: ... \\ \\
\medskip
{\normalfont\bfseries OUTPUT FORMAT:}

rewritten\_query: <the rewritten query>

\end{promptbox2}

\subsection*{GEPA Reflection Prompt}
\label{prompt:gepa_feedback}
\begin{promptbox2}
{\ttfamily\small

{\normalfont\bfseries INPUT:}

Ground-truth Lean statement: ... \\
Description of the Lean statement: ... \\ \\

Original query: ...\\ 
Original query ground-truth rank: ...\\ \\

Current rewritten query: ...\\
Current rewritten query ground-truth rank: ...\\ \\

Seed-prompt rewritten query: ...\\
Seed-prompt rewritten query ground truth rank: ...\\ \\

Optimization score: ...\\ \\
Retrieval outcome: ... \\

...

\medskip
{\normalfont\bfseries OBJECTIVE:}

Improve the reusable rewriting instruction ...

\medskip
{\normalfont\bfseries REFLECTION PRINCIPLES:}

1. Use the provided examples to diagnose rewriting behavior, not to memorize their answers.

\smallskip
2. Compare the current, original, and seed outcomes when available.
Consider both gains and regressions, distinguish observations from causal
hypotheses, and propose a change only when its applicability follows from
the query itself and is likely to transfer to unrelated topics.

\smallskip
3. Consider where a proposed change could hurt, and do not invent supporting evidence.

\smallskip
4. Remove inherited example-specific rules. Do not add topic or vocabulary
lists, target names, copied formulas, worked examples, or query-to-target mappings.

...\\

}
\end{promptbox2}

\subsection*{Query Generator Prompt}
\begin{promptbox2}
\ttfamily\small
You are a mathematician experienced in Lean4 helping stress-test a semantic search engine over Mathlib4.
Given the following Mathlib4 statement, generate a single search query that is genuinely relevant to the statement, while making the query adversarial to the retriever so it struggles to rank this statement highly.

\medskip
{\normalfont\bfseries Guidelines:}

- Relevance is mandatory: a mathematician reading the query and the statement must agree the statement is a correct answer.

\smallskip
- Make it something that a mathematician might actually write when searching for this statement.

\smallskip
- The query should not contain the formal name.

\smallskip
- The query should not be a near-verbatim rephrasing of the informal description.

\smallskip
- Keep it concise.

\smallskip
...

\medskip
Lean4 Statement:
...
\\
\\
Informal Description:
...

\medskip
Write your query wrapped in <search> and </search> tags. Output nothing else.
\end{promptbox2}

\end{document}